# Pac-Learning Recursive Logic Programs: Efficient Algorithms

**William W. Cohen**                                    WCOHEN@RESEARCH.ATT.COM
*AT&T Bell Laboratories*
*600 Mountain Avenue, Murray Hill, NJ 07974 USA*

## Abstract

We present algorithms that learn certain classes of function-free recursive logic programs in polynomial time from equivalence queries. In particular, we show that a single $k$-ary recursive constant-depth determinate clause is learnable. Two-clause programs consisting of one learnable recursive clause and one constant-depth determinate non-recursive clause are also learnable, if an additional "basecase" oracle is assumed. These results immediately imply the pac-learnability of these classes. Although these classes of learnable recursive programs are very constrained, it is shown in a companion paper that they are maximally general, in that generalizing either class in any natural way leads to a computationally difficult learning problem. Thus, taken together with its companion paper, this paper establishes a boundary of efficient learnability for recursive logic programs.

## 1. Introduction

One active area of research in machine learning is learning concepts expressed in first-order logic. Since most researchers have used some variant of Prolog to represent learned concepts, this subarea is sometimes called *inductive logic programming (ILP)* (Muggleton, 1992; Muggleton & De Raedt, 1994).

Within ILP, researchers have considered two broad classes of learning problems. The first class of problems, which we will call here *logic based relational learning* problems, are first-order variants of the sorts of classification problems typically considered within AI machine learning community: prototypical examples include Muggleton *et al.*'s (1992) formulation of $\alpha$-helix prediction, King *et al.*'s (1992) formulation of predicting drug activity, and Zelle and Mooney's (1994) use of ILP techniques to learn control heuristics for deterministic parsers. Logic-based relational learning often involves noisy examples that reflect a relatively complex underlying relationship; it is a natural extension of propositional machine learning, and has already enjoyed a number of experimental successes.

In the second class of problems studied by ILP researchers, the target concept is a Prolog program that implements some common list-processing or arithmetic function; prototypical problems from this class might be learning to append two lists, or to multiply two numbers. These learning problems are similar in character to those studied in the area of automatic programming from examples (Summers, 1977; Biermann, 1978), and hence might be appropriately called *automatic logic programming* problems. Automatic logic programming problems are characterized by noise-free training data and recursive target concepts. Thus a problem that is central to the enterprise of automatic logic programming—but not, perhaps, logic-based relational learning—is the problem of learning *recursive* logic programs.

The goal of this paper is to formally analyze the learnability of recursive logic programs in Valiant's (1984) model of pac-learnability, thus hopefully shedding some light on the task of automatic logic programming. To summarize our results, we will show that some simple recursive programs are pac-learnable from examples alone, or from examples plus a small number of additional "hints". The largest learnable class we identify in a standard learning model is the class of one-clause constant-depth determinate programs with at most a constant number of "closed" recursive literals. The largest learnable class we identify that requires extra "hints" is the class of constant-depth determinate programs consisting of a single nonrecursive base clause and a single recursive clause from the class described above. All of our results are proved in the model of *identification from equivalence queries* (Angluin, 1988, 1989), which is somewhat stronger than pac-learnability. Identification from equivalence queries requires that the target concept be exactly identified, in polynomial time, and using only a polynomial number of *equivalence queries*. An equivalence query asks if a hypothesis program $H$ is equivalent to the target program $C$; the answer to a query is either "yes" or an adversarily chosen example on which $H$ and $C$ differ. This model of learnability is arguably more appropriate for automatic logic programming tasks than the weaker model of pac-learnability, as it is unclear how often an approximately correct recursive program will be useful.

Interestingly, the learning algorithms analyzed are different from most existing ILP learning methods; they all employ an unusual method of generalizing examples called *forced simulation*. Forced simulation is a simple and analytically tractable alternative to other methods for generalizing recursive programs against examples, such as $n$-th root finding (Muggleton, 1994), sub-unification (Aha, Lapointe, Ling, & Matwin, 1994) and recursive anti-unification (Idestam-Almquist, 1993), but it has been only rarely used in experimental ILP systems (Ling, 1991).

The paper is organized as follows. After presenting some preliminary definitions, we begin by presenting (primarily for pedagogical reasons) a procedure for identifying from equivalence queries a single *non-recursive* constant-depth determinate clause. Then, in Section 4, we extend this learning algorithm, and the corresponding proof of correctness, to a simple class of recursive clauses: the class of "closed" linear recursive constant-depth determinate clauses. In Section 5, we relax some assumptions made to make the analysis easier, and present several extensions to this algorithm: we extend the algorithm from linear recursion to $k$-ary recursion, and also show how a $k$-ary recursive clause and a non-recursive clause can be learned simultaneously given an additional "basecase" oracle. We then discuss related work and conclude.

Although the learnable class of programs is large enough to include some well-known automatic logic programming benchmarks, it is extremely restricted. In a companion paper (Cohen, 1995), we provide a number of negative results, showing that relaxing any of these restrictions leads to difficult learning problems: in particular, learning problems that are either as hard as learning DNF (an open problem in computational learning theory), or as hard as cracking certain presumably secure cryptographic schemes. Thus, taken together with the results of the companion paper, our results delineate a boundary of learnability for recursive logic programs.

Although the two papers are independent, we suggest that readers wishing to read both this paper and the companion paper read this paper first.

## 2. Background

In this section we will present the technical background necessary to state our results. We will assume, however, that the reader is familiar with the basic elements of logic programming; readers without this background are referred to one of the standard texts, for example (Lloyd, 1987).

### 2.1 Logic Programs

Our treatment of logic programs is standard, except that we will usually consider the body of a clause to be an ordered set of literals.

For most of this paper, we will consider logic programs without function symbols—*i.e.*, programs written in Datalog.[1] The purpose of such a logic program is to answer certain questions relative to a *database*, $DB$, which is a set of ground atomic facts. (When convenient, we will also think of $DB$ as a conjunction of ground unit clauses.) The simplest use of a Datalog program is to check the status of a *simple instance*. A *simple instance* (for a program $P$ and a database $DB$) is a fact $f$. The pair $(P, DB)$ is said to *cover* $f$ iff $DB \wedge P \vdash f$. The set of simple instances covered by $(P, DB)$ is precisely the minimal model of the logic program $P \wedge DB$.

In this paper, we will primarily consider *extended instances* which consist of two parts: an *instance fact* $f$, which is simply a ground fact, and a *description* $D$, which is a finite set of ground unit clauses. An extended instance $e = (f, D)$ is covered by $(P, DB)$ iff

$$DB \wedge D \wedge P \vdash f$$

If extended instances are allowed, then function-free programs are expressive enough to encode surprisingly interesting programs. In particular, many programs that are usually written with function symbols can be re-written as function-free programs, as the example below illustrates.

**Example.** Consider the usual program for appending two lists.

append([],Ys,Ys).
append([X|Xs1],Ys,[X|Zs1]) ← append(Xs1,Ys,Zs1).

One could use this program to classify atomic facts containing function symbols such as *append([1,2],[3],[1,2,3])*. This program can be rewritten as a Datalog program that classifies extended instances as follows:

**Program $P$:**
append(Xs,Ys,Ys) ←
    null(Xs).
append(Xs,Ys,Zs) ←
    components(Xs,X,Xs1) ∧
    components(Zs,X,Zs1) ∧

---

1. This assumption is made primarily for convenience. In Section 5.2 we describe how this assumption can be relaxed.

append(Xs1,Ys,Zs1).

**Database** $DB$**:**
null(nil).

The predicate *components(A,B,C)* means that $A$ is a list with head $B$ and tail $C$; thus an extended instance equivalent to *append([1,2],[3],[1,2,3])* would be

**Instance fact** $f$**:**
append(list12,list3,list123).

**Description** $D$**:**
components(list12,1,list2).
components(list2,2,nil).
components(list123,1,list23).
components(list23,2,list3).
components(list3,3,nil).

We note that using extended instances as examples is closely related to using ground clauses entailed by the target clause as examples: specifically, the instance $e = (f, D)$ is covered by $P, DB$ iff $P \wedge DB \vdash (f \leftarrow D)$. As the example above shows, there is also a close relationship between extended instances and literals with function symbols that have been removed by "flattening" (Rouveirol, 1994; De Raedt & Džeroski, 1994). We have elected to use Datalog programs and the model of extended instances in this paper for several reasons. Datalog is relatively easy to analyze. There is a close connection between Datalog and the restrictions imposed by certain practical learning systems, such FOIL (Quinlan, 1990; Quinlan & Cameron-Jones, 1993), FOCL (Pazzani & Kibler, 1992), and GOLEM (Muggleton & Feng, 1992).

Finally, using extended instances addresses the following technical problem. The learning problems considered in this paper involve restricted classes of logic programs. Often, the restrictions imply that the number of simple instances is polynomial; we note that with only a polynomial-size domain, questions about pac-learnability are usually trivial. Requiring learning algorithms to work over the domain of extended instances precludes trivial learning techniques, however, as the number of extended instances of size $n$ is exponential in $n$ even for highly restricted programs.

## 2.2 Restrictions on Logic Programs

In this paper, we will consider the learnability of various restricted classes of logic programs. Below we will define some of these restrictions; however, we will first introduce some terminology.

If $A \leftarrow B_1 \wedge \ldots \wedge B_r$ is an (ordered) definite clause, then the *input variables* of the literal $B_i$ are those variables appearing in $B_i$ which also appear in the clause $A \leftarrow B_1 \wedge \ldots \wedge B_{i-1}$; all other variables appearing in $B_i$ are called *output variables*. Also, if $A \leftarrow B_1 \wedge \ldots \wedge B_r$ is a definite clause, then $B_i$ is said to be a *recursive literal* if it has the same predicate symbol and arity as $A$, the head of the clause.

### 2.2.1 Types of Recursion

The first set of restrictions concern the type of recursion that is allowed in a program. If every clause in a program has at most one recursive literal, then the program is *linear recursive*. If every clause in a program has at most $k$ recursive literals, then the program is *k-ary recursive*. Finally, if every recursive literal in a program contains no output variables, then we will say that the program is *closed recursive*.

### 2.2.2 Determinacy and Depth

The second set of restrictions are variants of restrictions originally introduced by Muggleton and Feng (1992). If $A \leftarrow B_1 \wedge \ldots \wedge B_r$ is an (ordered) definite clause, the literal $B_i$ is *determinate* iff for every possible substitution $\sigma$ that unifies $A$ with some fact $e$ such that

$$DB \vdash B_1 \sigma \wedge \ldots \wedge B_{i-1} \sigma$$

there is at most one maximal substitution $\theta$ so that $DB \vdash B_i \sigma \theta$. A clause is *determinate* if all of its literals are determinate. Informally, determinate clauses are those that can be evaluated without backtracking by a Prolog interpreter.

We also define the *depth* of a variable appearing in a clause $A \leftarrow B_1 \wedge \ldots \wedge B_r$ as follows. Variables appearing in the head of a clause have depth zero. Otherwise, let $B_i$ be the first literal containing the variable $V$, and let $d$ be the maximal depth of the input variables of $B_i$; then the depth of $V$ is $d + 1$. The depth of a clause is the maximal depth of any variable in the clause.

Muggleton and Feng define a logic program to be $ij$-determinate if it is is determinate, of constant depth $i$, and contains literals of arity $j$ or less. In this paper we use the phrase "constant-depth determinate" instead to denote this class of programs. Below are some examples of constant-depth determinate programs, taken from Džeroski, Muggleton and Russell (1992).

> **Example.** Assuming *successor* is functional, the following program is determinate. The maximum depth of a variable is one, for the variable $C$ in the second clause, and hence the program is of depth one.

less_than(A,B) ← successor(A,B).
less_than(A,B) ← successor(A,C) ∧ less_than(C,B).

The following program, which computes $\begin{pmatrix} A \\ C \end{pmatrix}$, is determinate and of depth two.

choose(A,B,C) ←
    zero(B) ∧
    one(C).
choose(A,B,C) ←
    decrement(B,D) ∧
    decrement(A,E) ∧

> multiply(B,C,G) ∧
> divide(G,A,F) ∧
> choose(E,D,F).

The program GOLEM (Muggleton & Feng, 1992) learns constant-depth determinate programs, and related restrictions have been adopted by several other practical learning systems (Quinlan, 1991; Lavrač & Džeroski, 1992; Cohen, 1993c). The learnability of constant-depth determinate clauses has also received some formal study, which we will review in Section 6.

### 2.2.3 MODE CONSTRAINTS AND DECLARATIONS

We define the *mode* of a literal $L$ appearing in a clause $C$ to be a string $s$ such that the initial character of $s$ is the predicate symbol of $L$, and for $j > 1$ the $j$-th character of $s$ is a "+" if the $(j-1)$-th argument of $L$ is an input variable and a "−" if the $(j-1)$-th argument of $L$ is an output variable. (This definition coincides with the usual definition of Prolog modes only when all arguments to the head of a clause are inputs. This simplification is justified, however, as we are considering only how clauses behave in classifying extended instances, which are ground.) A *mode constraint* is simply a set of mode strings $R = \{s_1, \ldots, s_k\}$, and a clause $C$ is said to *satisfy* a mode constraint $R$ for $p$ if for every literal $L$ in the body of $C$, the mode of $L$ is in $R$.

> **Example.** In the following *append* program, every literal has been annotated with its mode.

> append(Xs,Ys,Ys) ←
>     null(Xs).                    % mode: null+
> append(Xs,Ys,Zs) ←
>     components(Xs,X,Xs1) ∧       % mode: components + − −
>     components(Zs,X,Zs1) ∧       % mode: components + + −
>     append(Xs1,Ys,Zs1).          % mode: append + ++

The clauses of this program satisfy the following mode constraint:

$$\{ \quad components + --, \quad components + +-, \quad components + -+,$$
$$components - ++, \quad components + ++, \quad null+$$
$$append + +-, \qquad append + -+,$$
$$append - ++, \qquad append + ++ \qquad \}$$

Mode constraints are commonly used in analyzing Prolog code; for instance, they are used in many Prolog compilers. We will sometimes use an alternative syntax for mode constraints that parallels the syntax used in most Prolog systems: for instance, we may write the mode constraint "*components* + − −" as "*components*(+, −, −)".

We define a *declaration* to be a tuple $(p, a', R)$ where $p$ is a predicate symbol, $a'$ is an integer, and $R$ is a mode constraint. We will say that a clause $C$ *satisfies* a declaration if the head of $C$ has arity $a'$ and predicate symbol $p$, and if for every literal $L$ in the body of $C$ the mode of $L$ appears in $R$.

## 2.3 A Model of Learnability

In this section, we will present our model of learnability. We will first review the necessary definitions for a standard learning model, the model of learning from equivalence queries (Angluin, 1988, 1989), and discuss its relationship to other learning models. We will then introduce an extension to this model which is necessary for analyzing ILP problems.

### 2.3.1 Identification From Equivalence Queries

Let $X$ be a set. We will call $X$ the *domain*, and call the elements of $X$ *instances*. Define a *concept* $C$ over $X$ to be a representation of some subset of $X$, and define a *language* Lang to be a set of concepts. In this paper, we will be rather casual about the distinction between a concept and the set it represents; when there is a risk of confusion we will refer to the set represented by a concept $C$ as the *extension of* $C$. Two concepts $C_1$ and $C_2$ with the same extension are said to be *(semantically) equivalent*.

Associated with $X$ and Lang are two *size complexity measures*, for which we will use the following notation:

- The size complexity of a concept $C \in$ Lang is written $\|C\|$.

- The size complexity of an instance $e \in X$ is written $\|e\|$.

- If $S$ is a set, $S_n$ stands for the set of all elements of $S$ of size complexity no greater than $n$. For instance, $X_n = \{e \in X : \|e\| \leq n\}$ and $\text{Lang}_n = \{C \in \text{Lang} : \|C\| \leq n\}$.

We will assume that all size measures are polynomially related to the number of bits needed to represent $C$ or $e$.

The first learning model that we consider is the model of *identification with equivalence queries*. The goal of the learner is to *identify* some unknown *target concept* $C \in$ Lang— that is, to construct some hypothesis $H \in$ Lang such that $H \equiv C$. Information about the target concept is gathered only through *equivalence queries*. The input to an *equivalence query for* $C$ is some hypothesis $H \in$ Lang. If $H \equiv C$, then the response to the query is "yes". Otherwise, the response to the query is an arbitrarily chosen *counterexample*—an instance $e$ that is in the symmetric difference of $C$ and $H$.

A deterministic algorithm Identify *identifies* Lang *from equivalence queries* iff for every $C \in$ Lang, whenever Identify is run (with an oracle answering equivalence queries for $C$) it eventually halts and outputs some $H \in$ Lang such that $H \equiv C$. Identify *polynomially identifies* Lang *from equivalence queries* iff there is a polynomial $poly(n_t, n_e)$ such that at any point in the execution of Identify the total running time is bounded by $poly(n_t, n_e)$, where $n_t = \|C\|$ and $n_e$ is the size of the largest counterexample seen so far, or 0 if no equivalence queries have been made.

### 2.3.2 Relation to Pac-Learnability

The model of identification from equivalence queries has been well-studied (Angluin, 1988, 1989). It is known that if a language is learnable in this model, then it is also learnable in Valiant's (1984) model of pac-learnability. (The basic idea behind this result is that an equivalence query for the hypothesis $H$ can be emulated by drawing a set of random examples of a certain size. If any of them is a counterexample to $H$, then one returns the found counterexample as the answer to the equivalence query. If no counterexamples are found, one can assume with high confidence that $H$ is approximately equivalent to the target concept.) Thus identification from equivalence queries is a strictly stronger model than pac-learnability.

Most existing positive results on the pac-learnability of logic programs rely on showing that every concept in the target language can be emulated by a boolean concept from some pac-learnable class (Džeroski et al., 1992; Cohen, 1994). While such results can be illuminating, they are also disappointing, since one of the motivations for considering first-order representations in the first place is that they allow one to express concepts that cannot be easily expressed in boolean logic. One advantage of studying the exact identification model and considering recursive programs is that it essentially precludes use of this sort of proof technique: while many recursive programs can be approximated by boolean functions over a fixed set of attributes, few can be be exactly emulated by boolean functions.

### 2.3.3 BACKGROUND KNOWLEDGE IN LEARNING

The framework described above is standard, and is one possible formalization of the usual situation in inductive concept learning, in which a user provides a set of examples (in this case counterexamples to queries) and the learning system attempts to find a useful hypothesis. However, in a typical ILP system, the setting is slightly different, as usually the user provides clues about the target concept in addition to the examples. In most ILP systems the user provides a database $DB$ of "background knowledge" in addition to a set of examples; in this paper, we will assume that the user also provides a declaration. To account for these additional inputs it is necessary to extend the framework described above to a setting where the learner accepts inputs other than training examples.

To formalize this, we introduce the following notion of a "language family". If LANG is a set of clauses, $DB$ is a database and $Dec$ is a declaration, we will define LANG$[DB, Dec]$ to be the set of all pairs $(C, DB)$ such that $C \in$ LANG and $C$ satisfies $Dec$. Semantically, such a pair will denote the set of all extended instances $(f, D)$ covered by $(C, DB)$. Next, if $\mathcal{DB}$ is a set of databases and $\mathcal{DEC}$ is a set of declarations, then define

$$\text{LANG}[\mathcal{DB}, \mathcal{DEC}] = \{\text{LANG}[DB, Dec] : DB \in \mathcal{DB} \text{ and } Dec \in \mathcal{DEC}\}$$

This set of languages is called a *language family*.

We will now extend the definition of identification from equivalence queries to language families as follows. A language family LANG$[\mathcal{DB}, \mathcal{DEC}]$ is *identifiable from equivalence queries* iff every language in the set is identifiable from equivalence queries. A language family LANG$[\mathcal{DB}, \mathcal{DEC}]$ is *uniformly identifiable from equivalence queries* iff there is a single algorithm IDENTIFY$(DB, Dec)$ that identifies any language LANG$[DB, Dec]$ in the family given $DB$ and $Dec$.

Uniform polynomial identifiability of a language family is defined analogously: LANG$[\mathcal{DB}, \mathcal{DEC}]$ is *uniformly polynomially identifiable from equivalence queries* iff there is a polynomial time algorithm IDENTIFY$(DB, Dec)$ that identifies any language LANG$[DB, Dec]$ in the family given $DB$ and $Dec$. Note that IDENTIFY must run in time polynomial in the size of the inputs $Dec$ and $DB$ as well as the target concept.

### 2.3.4 Restricted Types of Background Knowledge

We will now describe a number of restricted classes of databases and declarations.

One restriction which we will make throughout this paper is to assume that all of the predicates of interest are of bounded arity. We will use the notation $a\text{-}\mathcal{DB}$ for the set of all databases that contain only facts of arity $a$ or less, and the notation $a\text{-}\mathcal{DEC}$ for the set of all declarations $(p, a', R)$ such that every string $s \in R$ is of length $a + 1$ or less.

For technical reasons, it will often be convenient to assume that a database contains an *equality predicate*—that is, a predicate symbol *equal* such that $equal(t_i, t_i) \in DB$ for every constant $t_i$ appearing in $DB$, and $equal(t_i, t_j) \notin DB$ for any $t_i \neq t_j$. Similarly, we will often wish to assume that a declaration allows literals of the form $equal(X, Y)$, where $X$ and $Y$ are input variables. If $\mathcal{DB}$ (respectively $\mathcal{DEC}$) is any set of databases (declarations) we will use $\mathcal{DB}^=$ ($\mathcal{DEC}^=$) to denote the corresponding set, with the additional restriction that the database (declaration) must contain an equality predicate (respectively the mode $equal(+, +)$).

It will sometimes also be convenient to assume that a declaration $(p, a', R)$ allows only a single valid mode for each predicate: *i.e.*, that for each predicate $q$ there is in $R$ only a single mode constraint of the form $q\alpha$. Such a declaration will be called a *unique-mode* declaration. If $\mathcal{DEC}$ is any set of declarations we will use $\mathcal{DEC}^1$ to denote the corresponding set of declarations with the additional restriction that the declaration is unique-mode.

Finally, we note that in a typical setting, the facts that appear in a database $DB$ and descriptions $D$ of extended instances are not arbitrary: instead, they are representative of some "real" predicate (*e.g.*, the relationship of a list to its *components* in the example above). One way of formalizing this is assume that all facts will be drawn from some restricted set $\mathcal{F}$; using this assumption one can define the notion of a *determinate mode*. If $f = p(t_1, \ldots, t_k)$ is a fact with predicate symbol $p$ and $p\alpha$ is a mode, then define $inputs(f, p\alpha)$ to be the tuple $\langle t_{i_1}, \ldots, t_{i_k} \rangle$, where $i_1, \ldots, i_k$ are the indices of $\alpha$ containing a "+". Also define $outputs(f, p\alpha)$ to be the tuple $\langle t_{j_1}, \ldots, t_{j_l} \rangle$, where $j_1, \ldots, j_l$ are the indices of $\alpha$ containing a "−". A mode string $p\alpha$ for a predicate $p$ is *determinate for* $\mathcal{F}$ iff the relation

$$\{ \langle inputs(f, p\alpha), outputs(f, p\alpha) \rangle : f \in \mathcal{F} \}$$

is a function. Informally, a mode is determinate if the input positions of the facts in $\mathcal{F}$ functionally determine the output positions.

The set of all declarations containing only modes determinate for $\mathcal{F}$ will be denoted $\mathcal{DetDEC}_{\mathcal{F}}$. However, in this paper, the set $\mathcal{F}$ will be assumed to be fixed, and thus we will generally omit the subscript.

A program consistent with a determinate declaration $Dec \in \mathcal{DetDEC}$ must be determinate, as defined above; in other words, consistency with a determinate declaration is a sufficient condition for semantic determinacy. It is also a condition that can be verified with a simple syntactic test.

### 2.3.5 Size Measures for Logic Programs

Assuming that all predicates are arity $a$ or less for some constant $a$ also allows very simple size measures to be used. In this paper, we will measure the size of a database $DB$ by its cardinality; the size of an extended instance $(f, D)$ by the cardinality of $D$; the size of a

509

declaration $(p, a', R)$ by the cardinality of $R$; and the size of a clause $A \leftarrow B_1 \wedge \ldots \wedge B_r$ by the number of literals in its body.

## 3. Learning a Nonrecursive Clause

The learning algorithms presented in this paper all use a generalization technique which we call *forced simulation*. By way of an introduction to this technique, we will consider a learning algorithm for non-recursive constant-depth clauses. While this result is presented primarily for pedagogical reasons, it may be of interest on its own: it is independent of previous proofs of the pac-learnability of this class (Džeroski et al., 1992), and it is also somewhat more rigorous than previous proofs.

Although the details and analysis of the algorithm for non-recursive clauses are somewhat involved, the basic idea behind the algorithm is quite simple. First, a highly-specific "bottom clause" is constructed, using two operations that we call *DEEPEN* and *CONSTRAIN*. Second, this bottom clause is generalized by deleting literals so that it covers the positive examples: the algorithm for generalizing a clause to cover an example is (roughly) to simulate the clause on the example, and delete any literals that would cause the clause to fail. In the remainder of this section we will describe and analyze this learning algorithm in detail.

### 3.1 Constructing a "Bottom Clause"

Let $Dec = (p, a', R)$ be a declaration and let $A \leftarrow B_1 \wedge \ldots \wedge B_r$ be a definite clause. We define

$$DEEPEN_{Dec}(A \leftarrow B_1 \wedge \ldots \wedge B_r) \equiv A \leftarrow B_1 \wedge \ldots \wedge B_r \wedge (\bigwedge_{L_i \in \mathcal{L}_D} L_i)$$

where $\mathcal{L}_D$ is a maximal set of literals $L_i$ that satisfy the following conditions:

- the clause $A \leftarrow B_1 \wedge \ldots \wedge B_r \wedge L_i$ satisfies the mode constraints given in $R$;

- if $L_i \in \mathcal{L}_D$ has the same mode and predicate symbol as some other $L_j \in \mathcal{L}_D$, then the input variables of $L_i$ are different from the input variables of $L_j$;

- every $L_i$ has at least one output variable, and the output variables of $L_i$ are all different from each other, and are also difference from the output variables of any other $L_j \in \mathcal{L}_D$.

As an extension of this notation, we define $DEEPEN_{Dec}^i(C)$ to be the result of applying the function $DEEPEN_{Dec}$ repeatedly $i$ times to $C$, *i.e.*,

$$DEEPEN_{Dec}^i(C) \equiv \begin{cases} C & \text{if } i = 0 \\ DEEPEN_{Dec}(DEEPEN_{Dec}^{i-1}(C)) & \text{otherwise} \end{cases}$$

We define the function $CONSTRAIN_{Dec}$ as

$$CONSTRAIN_{Dec}(A \leftarrow B_1 \wedge \ldots \wedge B_r) \equiv A \leftarrow B_1 \wedge \ldots \wedge B_r \wedge (\bigwedge_{L_i \in \mathcal{L}_C} L_i)$$

where $\mathcal{L}_C$ is the set of all literals $L_i$ such that $A \leftarrow B_1 \wedge \ldots \wedge B_r \wedge L_i$ satisfies the mode constraints given in $R$, and $L_i$ contains no output variables.

**Example.** Let $D0$ be the declaration $(p, 2, R)$ where $R$ contains the mode constraints $mother(+, -)$, $father(+, -)$, $male(+)$, $female(+)$, and $equal(+, +)$. Then

$DEEPEN_{D0}(\text{p(X,Y)}\leftarrow) \equiv$
    p(X,Y)←mother(X,XM)∧father(X,XF)∧ mother(Y,YM)∧father(Y,YF)

$DEEPEN_{D0}^{2}(\text{p(X,Y)}\leftarrow) \equiv DEEPEN_{D0}(DEEPEN_{D0}(\text{p(X,Y)}\leftarrow)) \equiv$
    p(X,Y)←
        mother(X,XM)∧father(X,XF)∧ mother(Y,YM)∧father(Y,YF)∧
        mother(XM,XMM)∧father(XM,XMF)∧ mother(XF,XFM)∧father(XF,XFF)∧
        mother(YM,YMM)∧father(YM,YMF)∧ mother(YF,YFM)∧father(YF,YFF)

$CONSTRAIN_{D0}(DEEPEN_{D0}(\text{p(X,Y)}\leftarrow)) \equiv$
    p(X,Y)←
        mother(X,XM)∧father(X,XF)∧ mother(Y,YM)∧father(Y,YF)∧
        male(X)∧female(X)∧male(Y)∧female(Y)∧
        male(XM)∧female(XM)∧male(XF)∧female(XF)∧
        male(YM)∧female(YM)∧male(YF)∧female(YF)∧
        equal(X,X)∧equal(X,XM)∧equal(X,XF)∧
        equal(X,Y)∧equal(X,YM)∧equal(X,YF)∧
        equal(XM,X)∧equal(XM,XM)∧equal(XM,XF)∧
        equal(XM,Y)∧equal(XM,YM)∧equal(XM,YF)∧
        equal(XF,X)∧equal(XF,XM)∧equal(XF,XF)∧
        equal(XF,Y)∧equal(XF,YM)∧equal(XF,YF)∧
        equal(Y,X)∧equal(Y,XM)∧equal(Y,XF)∧
        equal(Y,Y)∧equal(Y,YM)∧equal(Y,YF)∧
        equal(YM,X)∧equal(YM,XM)∧equal(YM,XF)∧
        equal(YM,Y)∧equal(YM,YM)∧equal(YM,YF)∧
        equal(YF,X)∧equal(YF,XM)∧equal(YF,XF)∧
        equal(YF,Y)∧equal(YF,YM)∧equal(YF,YF)

Let us say that clause $C_1$ is a *subclause* of clause $C_2$ if the heads of $C_1$ and $C_2$ are identical, if every literal in the body of $C_1$ also appears in $C_2$, and if the literals in the body of $C_1$ appear in the same order as they do in $C_2$. The functions $DEEPEN$ and $CONSTRAIN$ allow one to easily describe a clause with an interesting property.

**Theorem 1** *Let $Dec = (p, a', R)$ be a declaration in $a\text{-}\mathcal{D}et\mathcal{DEC}^{=}$, let $X_1, \ldots, X_{a'}$ be distinct variables, and define the clause $BOTTOM_d^*$ as follows:*

$$BOTTOM_d^*(Dec) \equiv CONSTRAIN_{Dec}(DEEPEN_{Dec}^d(p(X_1, \ldots, X_{a'})\leftarrow))$$

*For any constants $d$ and $a$, the following are true:*

- *the size of $BOTTOM_d^*(Dec)$ is polynomial in $\|Dec\|$;*
- *every depth-$d$ clause that satisfies $Dec$ (and hence, is determinate) is (semantically) equivalent to some subclause of $BOTTOM_d^*(Dec)$.*

**begin algorithm** $Force1_{NR}(d, Dec, DB)$:
    % below $BOTTOM_d^*$ is the most specific possible clause
    **let** $H \leftarrow BOTTOM_d^*(Dec)$
    **repeat**
        $Ans \leftarrow$ answer to the query "Is $H$ correct?"
        **if** $Ans =$ "yes" **then return** $H$
        **elseif** $Ans$ is a negative example **then**
            **return** "no consistent hypothesis"
        **elseif** $Ans$ is a positive example $e^+$ **then**
            % generalize $H$ minimally to cover $e^+$
            **let** $(f, D)$ be the components of the extended instance $e^+$
            $H \leftarrow ForceSim_{NR}(H, f, Dec, (DB \cup D))$
            **if** $H = FAILURE$ **then**
                **return** "no consistent hypothesis"
            **endif**
        **endif**
    **endrepeat**
**end**

Figure 1: A learning algorithm for nonrecursive depth-$d$ determinate clauses

**Proof:** See Appendix A. A related result also appears in Muggleton and Feng (1992). ■

**Example.** Below $C_1$ and $D_1$ are equivalent, as are $C_2$ and $D_2$. Notice that $D_1$ and $D_2$ are subclauses of $BOTTOM_1^*(D0)$.

$C_1$ : p(A,B)←mother(A,C)∧father(A,D)∧ mother(B,C)∧father(B,D)∧male(A)
$D_1$ : p(X,Y)←mother(X,XM)∧father(X,XF)∧ mother(Y,YM)∧father(Y,YF)∧
        male(X)∧equal(XM,YM)∧equal(XF,YF)
$C_2$ : p(A,B)←father(A,B)∧female(A)
$D_2$ : p(X,Y)←father(X,XF)∧female(X)∧equal(XF,Y)

For $C_1$ and $D_1$, $p(X,Y)$ is true when $X$ is $Y$'s brother. For $C_2$ and $D_2$, $p(X,Y)$ is true when $X$ is $Y$'s daughter, and $Y$ is $X$'s father.

## 3.2 The Learning Algorithm

Theorem 1 suggests that it may be possible to learn non-recursive constant-depth determinate clauses by searching the space of subclauses of $BOTTOM_d^*$ in some efficient manner. Figures 1 and 2 present an algorithm called $Force1_{NR}$ that does this when $Dec$ is a unique-mode declaration.

Figure 1 presents the top-level learning algorithm, $Force1_{NR}$. $Force1_{NR}$ takes as input a database $DB$ and a declaration $Dec$, and begins by hypothesizing the clause $BOTTOM_d^*(Dec)$. After each positive counterexample $e^+$, the current hypothesis is generalized as little as possible in order to cover $e^+$. This strategy means that the hypothesis is **begin subroutine** $ForceSim_{NR}(H, f, Dec, DB)$:
    % "forcibly simulate" $H$ on fact $f$
    **if** $f \in DB$ **then return** $H$
    **elseif** the head of $H$ and $f$ cannot be unified **then**
        **return** $FAILURE$
    **else**
        **let** $H' \leftarrow H$
        **let** $\sigma$ be the mgu of $f$ and the head of $H'$
        **for** each literal $L$ in the body of $H'$ **do**
            **if** there is a substitution $\sigma'$ such that $L\sigma\sigma' \in DB$ **then**
                $\sigma \leftarrow \sigma \circ \sigma'$, where $\sigma'$ is the most general such substitution
            **else**
                delete $L$ from the body of $H'$, together with
                 all literals $L'$ supported (directly or indirectly) by $L$
            **endif**
        **endfor**
        **return** $H'$
    **endif**
**end**

Figure 2: Forced simulation for nonrecursive depth-$d$ determinate clauses

always the least general hypothesis that covers the positive examples; hence, if a negative counterexample $e^-$ is ever seen, the algorithm will abort with a message that no consistent hypothesis exists.

To minimally generalize a hypothesis $H$, the function $ForceSim_{NR}$ is used. This subroutine is shown in Figure 2. In the figure, the following terminology is used. If some output variable of $L$ is an input variable of $L'$, then we say that $L$ *directly supports* $L'$. We will say that $L$ *supports* $L'$ iff $L$ directly supports $L'$, or if $L$ directly supports some literal $L''$ that supports $L'$. (Thus "supports" is the transitive closure of "directly supports".) $ForceSim_{NR}$ deletes from $H$ the minimal number of literals necessary to let $H$ cover $e^+$. To do this, $ForceSim_{NR}$ simulates the action of a Prolog interpreter in evaluating $H$, except that whenever a literal $L$ in the body of $H$ would fail, that literal is deleted, along with all literals $L'$ supported by $L$.

The idea of learning by repeated generalization is an old one; in particular, previous methods exist for learning a definite clause by generalizing a highly-specific one. For example, CLINT (De Raedt & Bruynooghe, 1992) generalizes a "starting clause" guided by queries made to the user; PROGOL (Srinivasan, Muggleton, King, & Sternberg, 1994) guides a top-down generalization process with a known bottom clause; and Rouveirol (1994) describes a method for generalizing bottom clauses created by saturation. The $Force1_{NR}$ algorithm is thus of interest not for its novelty, but because it is provably correct and efficient, as noted in the theorem below.

In particular, let $d$-DepthNonRec be the language of nonrecursive clauses of depth $d$ or less (and hence $i$-DepthNonRec$[\mathcal{DB}, j\text{-}\mathcal{D}et\mathcal{DEC}]$ is the language of nonrecursive $ij$-determinate clauses). We have the following result:

**Theorem 2** *For any constants $a$ and $d$, the language family*

$$d\text{-}\text{DepthNonRec}[\mathcal{DB}^=, a\text{-}\mathcal{D}et\mathcal{DEC}^{=1}]$$

*is uniformly identifiable from equivalence queries.*

**Proof:** We will show that $Force1_{NR}$ uniformly identifies this language family with a polynomial number of queries. We begin with the following important lemma, which characterizes the behavior of $ForceSim_{NR}$.

**Lemma 3** *Let $Dec$ declaration in $\mathcal{D}et\mathcal{DEC}^{=1}$, let $DB$ be a database, let $f$ be a fact, and let $H$ be a determinate nonrecursive clause that satisfies $Dec$. Then one of following conditions must hold:*

- *$ForceSim_{NR}(H, f, Dec, DB)$ returns FAILURE, and no subclause $H'$ of $H$ satisfies both $Dec$ and the constraint $H' \land DB \vdash f$; or,*

- *$ForceSim_{NR}(H, f, Dec, DB)$ returns a clause $H'$, and $H'$ is the unique syntactically largest subclause of $H$ that satisfies both $Dec$ and the constraint $H' \land DB \vdash f$.*

**Proof of lemma:** To avoid repetition, we will refer to the syntactically maximal subclauses $H'$ of $H$ that satisfy both $Dec$ and the constraint $H' \land DB \vdash f$ as "admissible subclauses" in the proof below.

Clearly the lemma is true if $H$ or *FAILURE* is returned by $ForceSim_{NR}$. In the remaining cases the **for** loop of the algorithm is executed, and we must establish these two claims (under the assumptions that $A$ and $f$ unify, and that $f \notin DB$):

**Claim 1.** If $L$ is retained, then every admissible subclause contains $L$.

**Claim 2.** If $L$ is deleted, then no admissible subclause contains $L$.

First, however, observe that deleting a literal $L$ may cause the mode of some other literals to violate the mode declarations of $Dec$. It is easy to see that if $L$ is deleted from a clause $C$, then the mode of all literals $L'$ directly supported by $L$ will change. Thus if $C$ satisfies a unique-mode declaration prior to the deletion of $L$, then after the deletion of $L$ all literals $L'$ that are directly supported by $L$ will have invalid modes.

Now, to see that Claim 1 is true, suppose instead that it is false. Then there must be some maximal subclause $C'$ of $H$ that satisfies $Dec$, covers the fact $f$, and does not contain $L$. By the argument above, if $C'$ does not contain $L$ but satisfied $Dec$, then $C'$ contains no literals $L'$ from $H$ that are supported by $L$. Hence the output variables of $L$ are disjoint from the variables appearing in $C'$. This means that if $L$ were to be added to $C'$ the resulting clause would still satisfy $Dec$ and cover $f$, which leads to a contradiction since $C'$ was assumed to be maximal.

To verify Claim 2, let us introduce the following terminology. If $C = (A \leftarrow B_1 \land \ldots \land B_r)$ is a clause and $DB$ is a database, we will say that the substitution $\theta$ is a $(DB, f)$-witness for $C$ iff $\theta$ is associated with a proof that $C \wedge DB \vdash f$ (or more precisely, iff $A\theta = f$ and $\forall i : 1 \leq i \leq r, B_i\theta \in DB$.) We claim that the following condition is an invariant of the **for** loop of the $ForceSim_{NR}$ algorithm.

**Invariant 1.** Let $C$ be any admissible subclause that contains all the literals in $H'$ preceding $L$ (*i.e.*, that contains all those literals of $H$ that were retained on previous iterations of the algorithm). Then every $(DB, f)$-witness for $C$ is a superset of $\sigma$.

This can be easily established by induction on the number of iterations of the *for* loop. The condition is true when the loop is first entered, since $\sigma$ is initially the most general unifier of $A$ and $f$. The condition remains true after an iteration in which $L$ is deleted, since $\sigma$ is unchanged. Finally, the condition remains true after an iteration in which $L$ is retained: because $\sigma'$ is maximally general, it may only assign values to the output variables of $L$, and by determinacy only one assignment to the output variables of $L$ can make $L$ true. Hence every $(DB, f)$-witness for $C$ must contain the bindings in $\sigma$.

Next, with an inductive argument and Claim 1 one can show that every admissible subclause $C$ must contain all the literals that have been retained in previous iterations of the loop, leading to the following strengthening of Invariant 1:

**Invariant 1'.** Let $C$ be any admissible subclause. Then every $(DB, f)$-witness for $C$ is a superset of $\sigma$.

Now, notice that only two types of literals are deleted: (a) literals $L$ such that no superset of $\sigma$ can make $L$ true, and (b) literals $L'$ that are supported by a literal $L$ of the preceding type. In case (a), clearly $L$ cannot be part of any admissible subclause, since no superset of $\sigma$ makes $L$ succeed, and only such supersets can be witnesses of admissible clauses. In case (b), again $L'$ cannot be part of any admissible subclause, since its declaration is invalid unless $L$ is present in the clause, and by the argument above $L$ cannot be in the clause.

This concludes the proof of the lemma. ∎

To prove the theorem, we must now establish the following properties of the identification algorithm.

**Correctness.** By Theorem 1, if the target program is in $d\text{-}\textsc{Depth\,Non\,Rec}[DB, Dec]$, then there is some clause $C_T$ that is equivalent to the target, and is a subclause of $BOTTOM_d^*(Dec)$. $H$ is initially $BOTTOM^*d$ and hence a superclause of $C_T$. Now consider invoking $ForceSim_{NR}$ on any positive counterexample $e^+$. By Lemma 3, if this invocation is successful, $H$ will be replaced by $H'$, the longest subclause of $H$ that covers $e^+$. Since $C_T$ is a subclause of $H$ that covers $e^+$, this means that $H'$ will again be a superclause of $C_T$. Inductively, then, the hypothesis is always a superclause of the target.

Further, since the counterexample $e^+$ is always an instance that is not covered by the current hypothesis $H$, every time the hypothesis is updated, the new hypothesis is a *proper* subclause of the old. This means that $Force1_{NR}$ will eventually identify the target clause.

**Efficiency.** The number of queries made is polynomial in $\|Dec\|$ and $\|DB\|$, since $H$ is initially of size polynomial in $\|Dec\|$, and is reduced in size each time a counterexample is provided. To see that each counterexample is processed in time polynomial in $n_r$, $n_e$, and $n_t$, notice that since the length of $H$ is polynomial, the number of repetitions of the **for** loop of $ForceSim_{NR}$ is also polynomial; further, since the arity of literals $L$ is bounded by

515

$a$, only $an_b + an_e$ constants exist in $DB \cup D$, and hence there are at most $(an_b + an_e)^a$ substitutions $\sigma'$ to check inside the **for** loop, which is again polynomial. Thus each execution of $ForceSim_{NR}$ requires only polynomial time.

This concludes the proof. ∎

## 4. Learning a Linear Closed Recursive Clause

Recall that if a clause has only one recursive literal, then the clause is *linear recursive*, and that if no recursive literal contains output variables, then the clause is *closed linear recursive*. In this section, we will describe how the *Force1* algorithm can be extended to learn a single linear closed recursive clause.[2] Before presenting the extension, however, we would first like to discuss a reasonable-sounding approach that, on closer examination, turns out to be incorrect.

### 4.1 A Remark on Recursive Clauses

One plausible first step toward extending *Force1* to recursive clauses is to allow recursive literals in hypotheses, and treat them the same way as other literals—that is, to include recursive literals in the initial clause $BOTTOM_d^*$, and delete these literals gradually as positives examples are received. A problem with this approach is that there is no simple way to check if a recursive literal in a clause succeeds or fails on a particular example. This makes it impossible to simply run $ForceSim_{NR}$ on clauses containing recursive literals.

A straightforward (apparent) solution to this problem is to assume that an oracle exists which can be queried as to the success or failure of any recursive literal. For closed recursive clauses, it is sufficient to assume that there is an oracle $\text{MEMBER}_{C_t}(DB, f)$ that answers the question

$$\text{Does } DB \wedge P \vdash f \text{ ?}$$

where $C_t$ is the unknown target concept, $f$ is a ground fact, and $DB$ is a database. Given such an oracle, one can determine if a closed recursive literal $L_r$ should be retained by checking if $\text{MEMBER}_{C_T}(DB, L_r\sigma)$ is true. Such an oracle is very close to the notion of a *membership query* as used in computational learning theory.

This is a natural extension of the $Force1_{NR}$ learning algorithm to recursive clauses—in fact an algorithm based on similar ideas has been been previously conjectured to pac-learn closed recursive constant-depth determinate clauses (Džeroski et al., 1992). Unfortunately, this algorithm can fail to return a clause that is consistent with a positive counterexample. To illustrate this, consider the following example.

**Example.** Consider using the extension of $Force1_{NR}$ described above to learn following target program:

append(Xs,Ys,Zs) ←

---

```
            components(Xs,X,Xs1),
            components(Zs,Z,Zs1),
            X1=Z1,
            append(Xs1,Ys,Zs1).
```

This program is determinate, has depth 1, and satisfies the following set of declarations:

```
components(+,−,−).
null(+).
equal(+,+).
odd(+).
append(+,+,+).
```

We will assume also a database $DB$ that defines the predicate *null* to be true for empty lists, and *odd* to be true for the constants 1 and 3.

To see how the forced simulation can fail, consider the following positive instance $e = (f, D)$:

$$f = append(l12, l3, l123)$$
$$D = \{ \ cons(l123,1,l23), \ cons(l23,2,l3), \ cons(l3,3,nil),$$
$$cons(l12,1,l2), \quad cons(l2,2,nil),$$
$$append(nil,l3,l3) \ \}$$

This is simply a "flattened" form of *append([1,2],[3],[1,2,3])*, together with the appropriate base case *append([],[3],[3])*. Now consider beginning with the clause $BOTTOM_1^*$ and generalizing it using $ForceSim_{NR}$ to cover this positive instance. This process is illustrated in Figure 3. The clause on the left in the figure is $BOTTOM_d^*(Dec)$; the clause on the right is the output of forcibly simulating this clause on $f$ with $ForceSim_{NR}$. (For clarity we've assumed that only the single correct recursive call remains after forced simulation.)

The resulting clause is incorrect, in that it does not cover the given example $e$. This can be easily seen by stepping through the actions of a Prolog interpreter with the generalized clause of Figure 3. The nonrecursive literals will all succeed, leading to the subgoal *append(l2,l3,l23)* (or in the usual Prolog notation, *append([2],[3],[2,3])*). This subgoal will fail at the literal *odd(X1)*, because *X1* is bound to 2 for this subgoal, and the fact *odd(2)* is not true in $DB \cup D$.

This example illustrates a pitfall in the policy of treating recursive and non-recursive literals in a uniform manner (For more discussion, see also (Bergadano & Gunetti, 1993; De Raedt, Lavrač, & Džeroski, 1993).) Unlike nonrecursive literals, the truth of the fact $L_r\sigma$ (corresponding to the recursive literal $L_r$) does not imply that a clause containing $L_r$ will succeed; it may be that while the first subgoal $L_r\sigma$ succeeds, deeper subgoals fail.

| $BOTTOM_1^*(Dec):$ | $ForceSim_{NR}(BOTTOM_1^*(Dec), f, Dec, DB \cup D):$ |
|---|---|
| append(Xs,Ys,Zs) ← | append(Xs,Ys,Zs) ← |
|     components(Xs,X1,Xs1)∧ |     components(Xs,X1,Xs1)∧ |
|     components(Ys,Y1,Ys1)∧ |     components(Ys,Y1,Ys1)∧ |
|     components(Zs,Z1,Zs1)∧ |     components(Zs,Z1,Zs1)∧ |
|     *null(Xs)∧* |     null(Ys1)∧ |
|     *null(Ys)∧* |     equal(X1,Z1)∧ |
|     ⋮ |     odd(X1)∧ |
|     null(Ys1)∧ |     odd(Y1)∧ |
|     *null(Zs1),* |     odd(Z1)∧ |
|     equal(Xs,Xs)∧ |     append(Xs1,Ys,Zs1). |
|     ⋮ | |
|     equal(X1,Z1)∧ | |
|     ⋮ | |
|     *equal(Zs1,Zs1)∧* | |
|     odd(Xs)∧ | |
|     ⋮ | |
|     odd(X1)∧ | |
|     odd(Y1)∧ | |
|     odd(Z1)∧ | |
|     ⋮ | |
|     *odd(Zs1)∧* | |
|     *append(Xs,Xs,Xs)∧* | |
|     ⋮ | |
|     *append(Zs1,Zs1,Zs1).* | |

Figure 3: A recursive clause before and after generalization with $ForceSim_{NR}$

## 4.2 Forced Simulation for Recursive Clauses

A solution to this problem is to replace the calls to the membership oracle in the algorithm sketched above with a call to a routine that forcibly simulates the actions of a top-down theorem-prover on a recursive clause. In particular, the following algorithm is suggested. First, build a nonrecursive "bottom clause", as was done in $ForceSim_{NR}$. Second, find some recursive literal $L_r$ such that appending $L_r$ to the bottom clause yields a recursive clause that can be generalized to cover the positive examples.

As in the nonrecursive case, a clause is generalized by deleting literals, using a straight-forward generalization of the procedure for forced simulation of nonrecursive clauses. During forced simulation, any failing nonrecursive subgoals are simply deleted; however, when a recursive literal $L_r$ is encountered, one forcibly simulates the hypothesis clause recursively **begin subroutine** $ForceSim(H, f, Dec, DB, h)$:
    *% "forcibly simulate" recursive clause $H$ on $f$*
    *% 1. check for infinite loops*
    **if** $h < 0$ **then return** $FAILURE$
    *% 2. check to see if $f$ is already covered*
    **elseif** $f \in DB$ **then return** $H$
    *% 3. check to see if $f$ cannot be covered*
    **elseif** the head of $H$ and $f$ cannot be unified **then**
        **return** $FAILURE$
    **else**
        **let** $L_r$ be the recursive literal of $H$
        **let** $H' \leftarrow H - \{L_r\}$
        *% 4. delete failing non-recursive literals as in $ForceSim_{NR}$*
        **let** $A$ be the head of $H'$
        **let** $\sigma$ be the mgu of $A$ and $e$
        **for** each literal $L$ in the body of $H'$ **do**
            **if** there is a substitution $\sigma'$ such that $L\sigma\sigma' \in DB$
            **then** $\sigma \leftarrow \sigma \circ \sigma'$, where $\sigma'$ is the most general such substitution
            **else**
                delete $L$ from the body of $H'$, together with
                 all literals $L'$ supported (directly or indirectly) by $L$
            **endif**
        **endfor**
        *% 5. generalize $H'$ on the recursive subgoal $L_r\sigma$*
        **if** $L_r\sigma$ is ground **then return** $ForceSim(H' \cup \{L_r\}, L_r\sigma, Dec, DB, h-1)$
        **else return** $FAILURE$
        **endif**
    **endif**
**end**

Figure 4: Forced simulation for linear closed recursive clauses

on the corresponding recursive subgoal. An implementation of forced simulation for linear closed recursive clauses is shown in Figure 4.

The extended algorithm is similar to $ForceSim_{NR}$, but differs in that when the recursive literal $L_r$ is reached in the simulation of $H$, the corresponding subgoal $L_r\sigma$ is created, and the hypothesized clause is recursively forcibly simulated on this subgoal. This ensures that the generalized clause will also succeed on the subgoal. For reasons that will become clear shortly, we would like this algorithm to terminate, even if the original clause $H$ enters an infinite loop when used in a top-down interpreter. In order to ensure termination, an extra argument $h$ is passed to $ForceSim$. The argument $h$ represents a depth bound for the forced simulation.

To summarize, the basic idea behind the algorithm of Figure 4 is to simulate the hypothesized clause $H$ on $f$, and generalize $H$ by deleting literals whenever $H$ would fail on $f$ or on any subgoal of $f$.

**Example.**

Consider using $ForceSim$ to forcibly simulate the following recursive clause $BOTTOM_1^*(Dec) \cup L_r$

append(Xs,Ys,Zs) ←
    components(Xs,X1,Xs1)∧components(Ys,Y1,Ys1)∧components(Zs,Z1,Zs1)∧
    null(Xs)∧...∧null(Zs1)∧
    odd(Xs)∧...∧odd(Zs1)∧
    equal(Xs,Xs)∧...∧equal(Zs1,Zs1)∧
    append(Xs1,Ys,Zs1)

Here the recursive literal $L_r$ is *append(Xs1,Ys,Zs1)*. We will also assume that $f$ is taken from the extended query $e = (f, D)$, which is again the flattened version of the instance *append([1,2],[3],[1,2,3])* used in the previous example; that $Dec$ is the set of declarations of in the previous example; and that the database $DB$ is $D \cup null(nul)$.

After executing steps 1-4 of $ForceSim$, a number of failing literals are deleted, leading to the substitution[3] $\sigma$ of $\{Xs = [1,2],\ Ys = [3],\ Zs = [1,2,3],\ X1 = 1,\ Xs1 = [2],\ Y1 = 3,\ Ys1 = [],\ Z1 = 1,\ Zs1 = [2,3]\}$ and the following reduced clause:

append(Xs,Ys,Zs) ←
    components(Xs,X1,Xs1)∧components(Ys,Y1,Ys1)∧components(Zs,Z1,Zs1)∧
    null(Ys1)∧odd(X1)∧odd(Y1)∧odd(Z1)∧equal(X1,Z1)∧
    append(Xs1,Ys,Zs1)

Hence the recursive subgoal is

$$L_r\sigma = append(Xs1,\ Ys,\ Zs1)\sigma = append([2],[3],[2,3])$$

---

3. Note that for readability, we are using the term notation rather than the flattened notation of $Xs = l12$, $Ys = l3$, etc.

Recursively applying *ForceSim* to this goal produces the substitution $\{Xs = [2],$
$Ys = [3], Zs = [2,3], X1 = 2, Xs1 = [], Y1 = 3, Ys1 = [], Z1 = 2, Zs1 = [3]\}$
and also results in deleting the additional literals *odd(X1)* and *odd(Z1)*. The
next recursive subgoal is $L_r\sigma = append([],[3],[3])$; since this clause is included
in the database $DB$, *ForceSim* will terminate. The final clause returned by
*ForceSim* in this case is the following:

append(Xs,Ys,Zs) ←
    components(Xs,X1,Xs1)∧components(Ys,Y1,Ys1)∧components(Zs,Z1,Zs1)∧
    null(Ys1)∧odd(Y1)∧equal(X1,Z1)∧
    append(Xs1,Ys,Zs1)

Notice that this clause does cover $e$.

As in Section 3 we begin our analysis by showing the correctness of the forced simulation
algorithm—*i.e.*, by showing that forced simulation does indeed produce a unique maximally
specific generalization of the input clause that covers the example.

This proof of correctness uses induction on the depth of a proof. Let us introduce again
some additional notation, and write $P \wedge DB \vdash_h f$ if the Prolog program $(P, DB)$ can be
used to prove the fact $f$ in a proof of depth $h$ or less. (The notion of depth of a proof is the
usual one; we will define looking up $f$ in the database $DB$ to be a proof of depth zero.) We
have the following result concerning the *ForceSim* algorithm.

**Theorem 4** *Let Dec be a declaration in $\mathcal{D}et\mathcal{DEC}^{=1}$, let $DB$ be a database, let $f$ be a fact,
and let $H$ be a determinate closed linear recursive clause that satisfies Dec. Then one of
the following conditions must hold:*

- *ForceSim$(H, f, Dec, DB, h)$ returns FAILURE, and no recursive subclause $H'$ of $H$
  satisfies both Dec and the constraint $H' \wedge DB \vdash_h f$; or,*

- *ForceSim$(H, f, Dec, DB, h)$ returns a clause $H'$, and $H'$ is the unique syntactically
  largest recursive subclause of $H$ that satisfies both Dec and the constraint $H' \wedge DB \vdash_h f$.*

**Proof:** Again to avoid repetition, we will refer to syntactically maximal recursive (non-
recursive) subclauses $H'$ of $H$ that satisfy both *Dec* and the constraint $H' \wedge DB \vdash_h f$ as
"admissible recursive (nonrecursive) subclauses" respectively.

The proof largely parallels the proof of Lemma 3—in particular, similar arguments
show that the clause returned by *ForceSim* satisfies the conditions of the theorem whenever
*FAILURE* is returned and whenever $H$ is returned. Note that the correctness of *ForceSim*
when $H$ is returned establishes the base case of the theorem for $h = 0$.

For the case of depth $h > 0$, let us assume the theorem holds for depth $h - 1$, and
proceed using mathematical induction. The arguments of Lemma 3 show that the following
condition is true after the **for** loop terminates.

**Invariant** $1'$. $H'$ is the unique maximal nonrecursive admissible subclause of $H$, and every
    $(DB, f)$-witness for $H'$ is a superset of $\sigma$.

**begin algorithm** *Force1*($d, Dec, DB$):

    % *below* $BOTTOM_d^*$ *is the most specific possible clause*

    **let** $L_{r_1}, \ldots, L_{r_p}$ be all possible closed recursive literals for $BOTTOM_d^*(Dec)$

    choose an unmarked recursive literal $L_{r_i}$

    **let** $H \leftarrow BOTTOM_d^*(Dec) \cup \{L_{r_i}\}$

    **repeat**

        $Ans \leftarrow$ answer to the query "Is $H$ correct?"

        **if** $Ans$ ="yes" **then return** $H$

        **elseif** $Ans$ is a negative example $e^-$ **then**

            $H \leftarrow FAILURE$

        **elseif** $Ans$ is a positive example $e^+$ **then**

            % *generalize* $H$ *minimally to cover* $e^+$

            **let** $(f, D)$ be the components of $e^+$

            $H \leftarrow ForceSim(H, f, Dec, (DB \cup D), (a\|D\| + a\|DB\|)^{a'})$

              where $a'$ is the arity of the clause head as given in $Dec$

        **endif**

        **if** $H = FAILURE$ **then**

            **if** all recursive literals are marked **then**

                **return** "no consistent hypothesis"

            **else**

                mark $L_{r_i}$

                choose an unmarked recursive literal $L_{r_j}$

                **let** $H \leftarrow BOTTOM_d^*(Dec) \cup \{L_{r_j}\}$

            **endif**

        **endif**

    **endrepeat**

**end**

Figure 5: A learning algorithm for nonrecursive depth-$d$ determinate clauses

Now, let us assume that there is some admissible recursive subclause $H^*$. Clearly $H^*$ must contain the recursive literal $L_r$ of $H$, since $L_r$ is the only recursive literal of $H$. Further, the nonrecursive clause $\hat{H} = H^* - \{L_r\}$ must certainly satisfy $Dec$ and also $\hat{H} \wedge DB \vdash f$, so it must (by the maximality of $H'$) be a subclause of $H'$. Hence $H^*$ must be a subclause of $H' \cup \{L_r\}$. Finally, if $L_r\sigma$ is ground (*i.e.*, if $L_r$ is closed in the clause $H' \cup L_r$) then by Invariant $1'$, the clause $H^*$ must also satisfy $H^* \wedge DB \vdash L_r\sigma$ by a proof of depth $h - 1$. (This is simply equivalent to saying that the recursive subgoal of $L_r\sigma$ generated in the proof must succeed.)

By the inductive hypothesis, then, the recursive call must return the unique maximal admissible recursive subclause of $H' \cup L_r$, which by the argument above must also be the unique maximal admissible recursive subclause of $H$.

Thus by induction the theorem holds. ∎

### 4.3 A Learning Algorithm for Linear Recursive Clauses

Given this method for generalizing recursive clauses, one can construct a learning algorithm for recursive clauses as follows. First, guess a recursive literal $L_r$, and make $H = BOTTOM_d^* \cup L_r$ the initial hypothesis of the learner. Then, ask a series of equivalence queries. After a positive counterexample $e^+$, use forced simulation to minimally generalize $H$ to cover $e^+$. After a negative example, choose another recursive literal $L'_r$, and reset the hypothesis to $H = BOTTOM_d^* \cup L'_r$.

Figure 5 presents an algorithm that operates along these lines. Let $d$-DepthLinRec denote the language of linear closed recursive clauses of depth $d$ or less. We have the following result:

**Theorem 5** *For any constants $a$ and $d$, the language family*

$$d\text{-DepthLinRec}[\mathcal{DB}^=, a\text{-}\mathcal{D}et\mathcal{DEC}^{=1}]$$

*is uniformly identifiable from equivalence queries.*

**Proof:** We will show that *Force1* uniformly identifies this language family with a polynomial number of queries.

**Correctness and query efficiency.** There are at most $a\|D\| + a\|DB\|$ constants in any set $DB \cup D$, at most $(a\|D\| + a\|DB\|)^{a'}$ $a'$-tuples of such constants, and hence at most $(a\|D\| + a\|DB\|)^{a'}$ distinct recursive subgoals $L_r\sigma$ that might be produced in proving that a linear recursive clause $C$ covers an extended instance $(f, D)$. Thus every terminating proof of a fact $f$ using a linear recursive clause $C$ must be of depth $(a\|D\| + a\|DB\|)^{a'}$ or less; *i.e.*, for $h = (a\|D\| + a\|DB\|)^{a'}$,

$$C \wedge DB \wedge D \vdash_h f \quad \text{iff} \quad C \wedge DB \wedge D \vdash f$$

Thus Theorem 4 can be strengthened: for the value of $h$ used in *Force1*, the subroutine *ForceSim* returns the syntactically largest subclause of $H$ that covers the example $(f, D)$ whenever any such a subclause exists, and returns *FAILURE* otherwise.

We now argue the correctness of the algorithm as follows. Assume that the hypothesized recursive literal is "correct"—*i.e.*, that the target clause $C_T$ is some subclause of $BOTTOM_d^* \cup L_r$. In this case it is easy to see that *Force1* will identify $C_T$, using an argument that parallels the one made for *Force1$_{NR}$*. Again by analogy to *Force1$_{NR}$*, it is easy to see that only a polynomial number of equivalence queries will be made involving the correct recursive literal.

Next assume that $L_r$ is not the correct recursive literal. Then $C_T$ need not be a subclause of $BOTTOM_d^* \cup L_r$, and the response to an equivalence query may be either a positive or negative counterexample. If a positive counterexample $e^+$ is received and *ForceSim* is called, then the result may be *FAILURE*, or it may be a proper subclause of $H$ that covers $e^+$. Thus the result of choosing an incorrect $L_r$ will be a (possibly empty) sequence of positive counterexamples followed by either a negative counterexample or *FAILURE*. Since all equivalence queries involving the correct recursive literal will be answered by either a positive counterexample or "yes"[4], then if a negative counterexample or *FAILURE* is obtained, it must be that $L_r$ is incorrect.

---

4. Recall that an answer of "yes" to an equivalence query means the hypothesis is correct.

The number of variables in $BOTTOM_d^*$ can be bounded by $a\|BOTTOM_d^*(Dec)\|$, and as each closed recursive literal is completely defined by an $a'$-tuple of variables, the number of possible closed recursive literals $L_r$ can be bounded by

$$p = (a\|BOTTOM_d^*(Dec)\|)^{a'}$$

Since $\|BOTTOM_d^*(Dec)\|$ is polynomial in $\|Dec\|$, $p$ is also polynomial in $\|Dec\|$. This means that only a polynomial number of incorrect $L_r$'s need to be discarded. Further since each successive hypothesis using a single incorrect $L_r$ is a proper subclause of the previous hypothesis, only a polynomial number of equivalence queries are needed to discard an incorrect $L_r$. Thus only a polynomial number of equivalence queries can be made involving incorrect recursive literals.

Thus *Force1* needs only a polynomial number of queries to identify $C_t$.

**Efficiency.** *ForceSim* runs in time polynomial in its arguments $H^*$, $f$, *Dec*, $DB \cup D$ and $h$. When *ForceSim* is called from *Force1*, $h$ is always polynomial in $n_e$ and $\|DB\|$, and $H$ is always no larger than $\|BOTTOM_d^*(Dec)\| + 1$, which in turn is polynomial in the size of *Dec*. Hence every invocation of *ForceSim* requires time polynomial in $n_e$, *Dec*, and $DB$, and hence *Force1* processes each query in polynomial time.

This completes the proof. ∎

This result is somewhat surprising, as it shows that recursive clauses can be learned even given an adversarial choice of training examples. In contrast, most implemented ILP systems require well-choosen examples to learn recursive clauses.

This formal result can also be strengthened in a number of technical ways. One of the more interesting strengthenings is to consider a variant of *Force1* that maintains a fixed set of positive and negative examples, and constructs the set of all least general clauses that are consistent with these examples: this could be done by taking each of the clauses $BOTTOM_d^* \cup L_{r_1}$, ..., $BOTTOM_d^* \cup L_{r_p}$, forcibly simulating them on each of the positive examples in turn, and then discarding those clauses that cover one or more negative examples. This set of clauses could then be used to tractably encode the version space of *all* consistent programs, using the $[S, N]$ representation for version spaces (Hirsh, 1992).

## 5. Extending the Learning Algorithm

We will now consider a number of ways in which the result of Theorem 5 can be extended.

### 5.1 The Equality-Predicate and Unique-Mode Assumptions

Theorem 5 shows that the language family

$$d\text{-}{\rm Depth}\,{\rm Lin}\,{\rm Rec}[\mathcal{DB}^=, a\text{-}\mathcal{D}et\mathcal{DEC}^{=1}]$$

is identifiable from equivalence queries. It is natural to ask if this result can be extended by dropping the assumptions that an equality predicate is present and that the declaration contains a unique legal mode for each predicate: that is, if the result can be extended to the language family

$$d\text{-}{\rm Depth}\,{\rm Lin}\,{\rm Rec}[\mathcal{DB}, a\text{-}\mathcal{D}et\mathcal{DEC}]$$

This extension is in fact straightforward. Given a database $DB$ and a declaration $Dec = (p, a', R)$ that do not satisfy the equality-predicate and unique-mode assumptions, one can modify them as follows.

1. For every constant $c$ appearing in $DB$, add the fact $equal(c, c)$ to $DB$.

2. For every predicate $q$ that has $k$ valid modes $qs_1, \ldots, qs_k$ in $R$:

    (a) remove the mode declarations for $q$, and replace them with $k$ mode strings for the $k$ new predicates $q_{s_1}, \ldots, q_{s_k}$, letting $q_{s_i} s_i$ be the unique legal mode for the predicate $q_{s_i}$;

    (b) remove every fact $q(t_1, \ldots, t_a)$ of the predicate $q$ from $DB$, and replace it with the $k$ facts $q_{s_1}(t_1, \ldots, t_a), \ldots, q_{s_k}(t_1, \ldots, t_a)$.

Note that if the arity of predicates is bounded by a constant $a$, then the number of modes $k$ for any predicate $q$ is bounded by the constant $2^a$, and hence these transformations can be performed in polynomial time, and with only a polynomial increase in the size of $Dec$ and $DB$.

Clearly any target clause $C_t \in d\text{-}\textsc{Depth}\textsc{Lin}\textsc{Rec}[DB, Dec]$ is equivalent to some clause $C_t' \in d\text{-}\textsc{Depth}\textsc{Lin}\textsc{Rec}[DB', Dec']$, where $DB'$ and $Dec'$ are the modified versions of $DB$ and $Dec$ constructed above. Using $Force1$ it is possible to identify $C_t'$. (In learning $C_t'$, one must also perform steps 1 and 2b above on the description part $D$ of every counterexample $(f, D)$.) Finally, one can convert $C_t'$ to an equivalent clause in $d\text{-}\textsc{Depth}\textsc{Lin}\textsc{Rec}[DB, Dec]$ by repeatedly resolving against the clause $equal(X, X) \leftarrow$, and also replacing every predicate symbol $q_{s_i}$ with $q$.

This leads to the following strengthening of Theorem 5:

**Proposition 6** *For any constants $a$ and $d$, the language family*

$$d\text{-}\textsc{Depth}\textsc{Lin}\textsc{Rec}[\mathcal{DB}, a\text{-}\mathcal{D}et\mathcal{DEC}]$$

*is uniformly identifiable from equivalence queries.*

## 5.2 The Datalog Assumption

So far we have assumed that the target program contains no function symbols, and that the background knowledge provided by the user is a database of ground facts. While convenient for formal analysis, these assumptions can be relaxed.

Examination of the learning algorithm shows that the database $DB$ is used in only two ways.

- In forcibly simulating a hypothesis on an extended instance $(f, D)$, it is necessary to find a substitution $\sigma'$ that makes a literal $L$ true in the database $DB \cup D$. While this can be done algorithmically if $DB$ and $D$ are sets of ground facts, it is also plausible to assume that the user has provided an oracle that answers in polynomial time any mode-correct query $L$ to the database $DB$. Specifically, the answer of the oracle will be either

- the (unique) most-general substitution $\sigma'$ such that $DB \wedge D \vdash L\sigma'$ and $L\sigma'$ is ground; or

- "no" if no such $\sigma'$ exists.

Such an oracle would presumably take the form of an efficient theorem-prover for $DB$.

- When calling *ForceSim*, the top-level learning algorithm uses $DB$ and $D$ to determine a depth bound on the length of a proof made using the hypothesis program. Again, it is reasonable to assume that the user can provide this information directly, in the form of an oracle. Specifically, this oracle would provide for any fact $f$ a polynomial upper bound on the depth of the proof for $f$ in the target program.

Finally we note that if efficient (but non-ground) background knowledge is allowed, then function symbols always can be removed via flattening (Rouveirol, 1994). This transformation also preserves determinacy, although it may increase depth—in general, the depth of a flattened clause depends also on term depth in the original clause. Thus, the assumption that the target program is in Datalog can be replaced by assumptions that the term depth is bounded by a constant, and that two oracles are available: an oracle that answers queries to the background knowledge, and a depth-bound oracle. Both types of oracles have been frequently assumed in the literature (Shapiro, 1982; Page & Frisch, 1992; Džeroski et al., 1992).

## 5.3 Learning $k$-ary Recursive Clauses

It is also natural to ask if Theorem 5 can be extended to clauses that are not linear recursive. One interesting case is the case of closed $k$-ary recursive clauses for constant $k$. It is straightforward to extend *Force1* to guess a tuple of $k$ recursive literals $L_{r_1}, \ldots, L_{r_k}$, and then to extend *ForceSim* to recursively generalize the hypothesis clause on each of the facts $L_{r_1}\sigma, \ldots, L_{r_k}\sigma$. The arguments of Theorems 4 and 5 can be modified to show that this extension will identify the target clause after a polynomial number of equivalence queries.

Unfortunately, however, it is no longer the case that *ForceSim* runs in polynomial time. This is easily seen if one considers a tree of all the recursive calls made by *ForceSim*; in general, this tree will have branching factor $k$ and polynomial depth, and hence exponential size. This result is unsurprising, as the implementation of *ForceSim* described forcibly simulates a depth-bounded top-down interpreter, and a $k$-ary recursive program can take exponential time to interpret with such an interpreter.

There are at least two possible solutions to this problem. One possible solution is to retain the simple top-down forced simulation procedure, and require the user to provide a depth bound tighter than $(a\|D\| + a\|DB\|)^{a'}$, the maximal possible depth of a tree. For example, in learning a 2-ary recursive sort such as quicksort, the user might specify a logarithmic depth bound, thus guaranteeing that *ForceSim* is polynomial-time. This requires additional input from the user, but would be easy to implement. It also has the advantage (not shared by the approach described below) that the hypothesized program can be executed using a simple depth-bounded Prolog interpreter, and will always have shallow proof trees. This seems to be a plausible bias to impose when learning $k$-ary recursive Prolog programs, as many of these tend to have shallow proof trees.

A second solution to the possible high cost of forced simulation for $k$-ary recursive programs is to forcibly simulate a "smarter" type of interpreter—one which can execute $k$-ary recursive program in polynomial time.[5] One sound and complete theorem-prover for closed $k$-ary recursive programs can be implemented as follows.

Construct a top-down proof tree in the usual fashion, *i.e.*, using a depth-first left-to-right strategy, but maintain a list of the ancestors of the current subgoal, and also a list *VISITED* that records, for each previously visited node in the tree, the subgoal associated with that node. Now, suppose that in the course of constructing the proof tree one generates a subgoal $f$ that is on the *VISITED* list. Since the traversal of the tree is depth-first left-to-right, the node associated with $f$ is either an ancestor of the current node, or is a descendant of some left sibling of an ancestor of the current node. In the former case, the proof tree contains a loop, and cannot produce a successful proof; in this case the theorem-prover should exit with failure. In the latter case, a proof must already exist for $f'$, and hence nodes below the current node in the tree need not be visited; instead the theorem prover can simply assume that $f$ is true.

This top-down interpreter can be easily extended into a forced simulation procedure: one simply traverses the tree in the same order, generalizing the current hypothesis $H$ as needed to justify each inference step in the tree. The only additional point to note is that if one is performing forced simulation and revisits a previously proved subgoal $f$ at a node $n$, the current clause $H$ need not be further generalized in order to prove $f$, and hence it is again permissible to simply skip the portion of the tree below $n$. We thus have the following result.

**Theorem 7** *Let $d$-Depth-$k$-Rec be the set of $k$-ary closed recursive clauses of depth $d$. For any constants $a$, $d$, and $k$ the language family*

$$d\text{-Depth-}k\text{-Rec}[\mathcal{DB}, a\text{-}\mathcal{D}et\mathcal{DEC}]$$

*is uniformly identifiable from equivalence queries.*

**Proof:** Omitted, but following the informal argument made above. ∎

Note that we give this result without the restrictions that the database contains an equality relation and that the declaration is unique-mode, since the tricks used to relax these restrictions in Proposition 6 are still applicable.

## 5.4 Learning Recursive and Base Cases Simultaneously

So far, we have analyzed the problem of learning single clauses: first a single nonrecursive clause, and then a single recursive clause. However, every useful recursive program contains at least two clauses: a recursive clause, and a nonrecursive base case. It is natural to ask if it is possible to learn a complete recursive program by simultaneously learning both a recursive clause, and its associated nonrecursive base case.

In general, this is not possible, as is demonstrated elsewhere (Cohen, 1995). However, there are several cases in which the positive result can be extended to two-clause programs.

---

5. Note that it is plausible to believe that such a theorem-prover exists, as there are only a polynomial number of possible theorem-proving goals—namely, the $(a\|D\| + a\|DB\|)^{a^d}$ possible recursive subgoals.

**begin algorithm** *Force2*($d$, *Dec*, *DB*):

    **let** $L_{r_1}, \ldots, L_{r_p}$ be all possible recursive literals for $BOTTOM_d^*(Dec)$

    choose an unmarked recursive literal $L_{r_i}$

    **let** $H_R^* \leftarrow BOTTOM_d^*(Dec) \cup \{L_{r_i}\}$

    **let** $H_B^* \leftarrow BOTTOM_d^*(Dec)$

    **let** $P = (H_R, H_b)$

    **repeat**

        $Ans \leftarrow$ answer to query "Is $H_R^*$, $H_B^*$ correct?"

        **if** $Ans =$ "yes" **then return** $H_R^*, H_B^*$

        **elseif** $Ans$ is a negative example $e^-$ **then**

            $P \leftarrow FAILURE$

        **elseif** $Ans$ is a positive example $e^+$ **then**

            **let** $(f, D)$ be the components of $e^+$

            $P \leftarrow ForceSim2(H_R^*, H_B^*, f, Dec, (DB \cup D), (a\|D\| + a\|DB\|)^{a'})$

        **endif**

        **if** $P = FAILURE$ **then**

            **if** all recursive literals $L_{r_j}$ are marked **then**

                **return** "no consistent hypothesis"

            **else**

                mark $L_{r_i}$

                choose an unmarked recursive literal $L_{r_j}$

                **let** $H_R^* \leftarrow BOTTOM_d^*(Dec) \cup \{L_{r_j}\}$

                **let** $H_R^* \leftarrow BOTTOM_d^*(Dec)$

                **let** $P = (H_R^*, H_B^*)$

            **endif**

        **endif**

    **endrepeat**

**end**

Figure 6: A learning algorithm for two-clause recursive programs

**begin subroutine** *ForceSim2*($H_R, H_B, f, Dec, DB, h$):
    *% "forcibly simulate" program $H_R, H_B$ on $f$*
    **if** $h < 1$ **then return** *FAILURE*
    *% check to see if $f$ should be covered by $H_B$*
    **elseif** *BASECASE*($f$) **then**
        *return current $H_r$ and generalized $H_B$*
        **return** ($H_R$, *ForceSim$_{NR}$*($H_B, f, Dec, DB$))
    **elseif** the head of $H_R$ and $f$ cannot be unified **then**
        **return** *FAILURE*
    **else**
        **let** $L_r$ be the recursive literal of $H_R$
        **let** $H' \leftarrow H - \{L_r\}$
        **let** $A$ be the head of $H'$
        **let** $\sigma$ be the mgu of $A$ and $e$
        **for** each literal $L$ in the body of $H'$ **do**
            **if** there is a substitution $\sigma'$ such that $L\sigma\sigma' \in DB$
            **then** $\sigma \leftarrow \sigma \circ \sigma'$, where $\sigma'$ is the most general such substitution
            **else**
                delete $L$ from the body of $H'$, together with
                all literals $L'$ supported (directly or indirectly) by $L$
            **endif**
        **endfor**
        *% generalize $H', H_B$ on the recursive subgoal $L_r\sigma$*
        **if** $L_r\sigma$ is ground **then**
            *% continue the simulation of the program*
            **return** *ForceSim2*($H' \cup \{L_r\}, H_B, L_r\sigma, Dec, DB, h - 1$)
        **else return** *FAILURE*
        **endif**
    **endif**
**end**

Figure 7: Forced simulation for two-clause recursive programs

In this section, we will first discuss learning a recursive clause and base clause simultaneously, assuming that any determinate base clause is possible, but also assuming that an additional "hint" is available, in the form of a special "basecase" oracle. We will then discuss various alternative types of "hints".

Let $P$ be a target program with base clause $C_B$ and recursive clause $C_R$. A *basecase oracle* for $P$ takes as input an extended instance $(f, D)$ and returns "yes" if $C_B \wedge DB \wedge D \vdash f$, and "no" otherwise. In other words, the oracle determines if $f$ is covered by the nonrecursive base clause alone. As an example, for the *append* program, the basecase oracle should return "yes" for an instance *append(Xs, Ys, Zs)* when $Xs$ is the empty list, and "no" otherwise.

Given the existence of a basecase oracle, the learning algorithm can be extended as follows. As before, all possible recursive literals $L_{r_i}$ of the clause $BOTTOM_d^*$ are generated; however, in this case, the learner will test two clause hypotheses that are initially of the form $(BOTTOM_d^* \cup L_{r_i}, BOTTOM_d^*)$. To forcibly simulate such a hypothesis on a fact $f$, the following procedure is used. After checking the usual termination conditions, the forced simulator checks to see if $BASECASE(f)$ is true. If so, it calls $ForceSim_{NR}$ (with appropriate arguments) to generalize the current hypothesis for the base case. If $BASECASE(f)$ is false, then the recursive clause $H_r$ is forcibly simulated on $f$, a subgoal $L_r\sigma$ is generated as in before, and the generalized program is recursively forcibly simulated on the subgoal. Figures 6 and 7 present a learning algorithm *Force2* for two clause programs consisting of one linear recursive clause $C_R$ and one nonrecursive clause $C_B$, under the assumption that both equivalence and basecase oracles are available.

It is straightforward to extend the arguments of Theorem 5 to this case, leading to the following result.

**Theorem 8** *Let d-*DEPTH-*2-*CLAUSE *be the set of 2-clause programs consisting of one clause in d-*DEPTHLINREC *and one clause in d-*DEPTHNONREC*. For any constants a and d the language family*

$$d\text{-}\text{DEPTH-2-CLAUSE}[\mathcal{DB}, a\text{-}\mathcal{D}et\mathcal{DEC}]$$

*is uniformly identifiable from equivalence and basecase queries.*

**Proof:** Omitted. ∎

A companion paper (Cohen, 1995) shows that something like the basecase oracle is necessary: in particular, without any "hints" about the base clause, learning a two-clause linear recursive program is as hard as learning boolean DNF. However, there are several situations in which the basecase oracle can be dispensed with.

**Case 1.** The basecase oracle can be replaced by a polynomial-sized set of possible base clauses. The learning algorithm in this case is to enumerate pairs of base clauses $C_{B_i}$ and "starting clauses" $BOTTOM^* \cup L_{r_j}$, generalize the starting clause with forced simulation, and mark a pair as incorrect if overgeneralization is detected.

**Case 2.** The basecase oracle can be replaced by a fixed rule that determines when the base clause is applicable. For example, consider the rule that says that the base clause is applicable to any atom $p(X_1, \ldots, X_a)$ such that no $X_i$ is a non-null list. Adopting such a rule leads immediately to a learning procedure that pac-learns exactly those two-clause linear recursive programs for which the rule is correct.

**Case 3.** The basecase oracle can be also be replaced by a polynomial-sized set of rules for determining when a base clause is applicable. The learning algorithm in this case is pick a unmarked decision rule and run *Force2* using that rule as a basecase oracle. If *Force2* returns "no consistent hypothesis" then the decision rule is marked incorrect, and a new one is choosen. This algorithm will learn those two-clause linear recursive programs for which any of the given decision rules is correct.

Even though the general problem of determining a basecase decision rule for an arbitrary Datalog program may be difficult, it may be that a small number of decision procedures apply to a large number of common Prolog programs. For example, the recursion for most list-manipulation programs halts when some argument is reduced to a null list or to a singleton list. Thus Case 3 above seems likely to cover a large fraction of the automatic logic programming programs of practical interest.

We also note that heuristics have been proposed for finding such basecase decision rules automatically using typing restrictions (Stahl, Tausend, & Wirth, 1993).

## 5.5 Combining the Results

Finally, we note that all of the extensions described above are compatible. This means that if we let $kd$-MaxRecLang be the language of two-clause programs consisting of one clause $C_R$ that is $k$-ary closed recursive and depth-$d$ determinate, and one clause $C_B$ that is nonrecursive and depth-$d$ determinate, then the following holds.

**Proposition 9** *For any constants a, k and d the language family*

$$kd\text{-MaxRecLang}[\mathcal{DB}, a\text{-}\mathcal{D}et\mathcal{DEC}]$$

*is uniformly identifiable from equivalence and basecase queries.*

### 5.5.1 Further Extensions

The notation $kd$-MaxRecLang may seem at this point to be unjustified; although it is the most expressive language of recursive clauses that we have proven to be learnable, there are numerous extensions that may be efficiently learnable. For example, one might generalize the language to allow an arbitrary number of recursive clauses, or to include clauses that are not determinate. These generalizations might very well be pac-learnable—given the results that we have presented so far.

However, a companion paper (Cohen, 1995) presents a series of negative results showing that most natural generalizations of $kd$-MaxRecLang are not efficiently learnable, and further that $kd$-MaxRecLang itself is not efficiently learnable without the basecase oracle. Specifically, the companion paper shows that eliminating the basecase oracle leads to a problem that is as hard as learning boolean DNF, an open problem in computational learning theory. Similarly, learning two linear recursive clauses simultaneously is as hard as learning DNF, even if the base case is known. Finally, the following learning problems are all as hard as breaking certain (presumably) secure cryptographic codes: learning $n$ linear recursive determinate clauses, learning one $n$-ary recursive determinate clause, or learning one linear recursive "$k$-local" clause. All of these negative results hold not only for the model of identification from equivalence queries, but also for the weaker models of pac-learnability and pac-predictability.

## 6. Related Work

In discussing related work we will concentrate on previous formal analyses that employ a learning model similar to that considered here: namely, models that (a) require all computation be polynomial in natural parameters of the problem, and (b) assume either a neutral source or adversarial source of examples, such as equivalence queries or stochastically presented examples. We note, however, that much previous formal work exists that relies on different assumptions. For instance, there has been much work in which member or subset queries are allowed (Shapiro, 1982; De Raedt & Bruynooghe, 1992), or where examples are chosen in some non-random manner that is helpful to the learner (Ling, 1992; De Raedt & Džeroski, 1994). There has also been some work in which the efficiency requirements imposed by the pac-learnability model are relaxed (Nienhuys-Cheng & Polman, 1994). If the requirement of efficiency is relaxed far enough, very general positive results can be obtained using very simple learning algorithms. For example, in model of learnability in the limit (Gold, 1967), any language that is both recursively enumerable and decidable (which includes all of Datalog) can be learned by a simple enumeration procedure; in the model of U-learnability (Muggleton & Page, 1994) any language that is polynomially enumerable and polynomially decidable can be learned by enumeration.

The most similar previous work is that of Frazier and Page (1993a, 1993b). They analyze the learnability from equivalence queries of recursive programs with function symbols but without background knowledge. The positive results they provide are for program classes that satisfy the following property: given a set of positive examples $S^+$ that requires all clauses in the target program to prove the instances in $S^+$, only a polynomial number of recursive clauses are possible; further the base clause must have a certain highly constrained form. Thus the concept class is "almost" bounded in size by a polynomial. The learning algorithm for such a program class is to interleave a series of equivalence queries that test every possible target program. In contrast, our positive results are for exponentially large classes of recursive clauses. Frazier and Page also present a series of negative results suggesting that the learnable languages that they analyzed are difficult to generalize without sacrificing efficient learnability.

Previous results also exist on the pac-learnability of nonrecursive constant-depth determinate programs, and on the pac-learnability of recursive constant-depth determinate programs in a model that also allows membership and subset queries (Džeroski et al., 1992).

The basis for the intelligent search used in our learning algorithms is the technique of *forced simulation*. This method finds the least implicant of a clause $C$ that covers an extended instance $e$. Although when we developed this method we believed it to be original, subsequently we discovered that this was not the case—an identical technique had been previously proposed by Ling (1991). Since an extended instance $e$ can be converted (via saturation) to a ground Horn clause, there is also a close connection between forced simulation and recent work on "inverting implication" and "recursive anti-unification"; for instance, Muggleton (1994) describes a nondeterministic procedure for finding all clauses that imply a clause $C$, and Idestam-Almquist (1993) describes a means of constraining such an implicant-generating procedure to produce the least common implicant of two clauses. However, while both of these techniques have obvious applications in learning, both are extremely expensive in the worst case.

The CRUSTACEAN system (Aha et al., 1994) uses inverting implication in constrained settings to learn certain restricted classes of recursive programs. The class of programs efficiently learned by this system is not formally well-understood, but it appears to be similar to the classes analyzed by Frazier and Page. Experimental results show that these systems perform well on inferring recursive programs that use function symbols in certain restricted ways. This system cannot, however, make use of background knowledge.

Finally, we wish to direct the reader to several pieces of our own research that are relevant. As noted above, a companion paper exists which presents negative learnability results for several natural generalizations of the language $kd$-MaxRecLang (Cohen, 1995). Another related paper investigates the learnability of non-recursive Prolog programs (Cohen, 1993b); this paper also contains a number of negative results which strongly motivate the restriction of constant-depth determinacy. A final prior paper which may be of interest presents some experimental results with a Prolog implementation of a variant of the *Force2* algorithm (Cohen, 1993a). This paper shows that forced simulation can be the basis of a learning program that outperforms state-of-the art heuristic methods such as FOIL (Quinlan, 1990; Quinlan & Cameron-Jones, 1993) in learning from randomly chosen examples.

## 7. Conclusions

Just as it is often desirable to have guarantees of correctness for a program, in many plausible contexts it would be highly desirable to have an automatic programming system offer some formal guarantees of correctness. The topic of this paper is the learnability of recursive logic programs using formally well-justified algorithms. More specifically, we have been concerned with the development of algorithms that are provably sound and efficient in learning recursive logic programs from equivalence queries. We showed that one constant-depth determinate closed $k$-ary recursive clause is identifiable from equivalent queries; this implies immediately that this language is also learnable in Valiant's (1984) model of pac-learnability. We also showed that a program consisting of one such recursive clause and one constant-depth determinate nonrecursive clause is identifiable from equivalence queries given an additional "basecase oracle", which determines if a positive example is covered by the non-recursive base clause of the target program alone.

In obtaining these results, we have introduced several new formal techniques for analyzing the learnability of recursive programs. We have also shown the soundness and efficiency of several instances of *generalization by forced simulation*. This method may have applications in practical learning systems. The *Force2* algorithm compares quite well experimentally with modern ILP systems on learning problems from the restricted class that it can identify (Cohen, 1993a); thus sound learning methods like *Force2* might be useful as a filter before a more general ILP system like FOIL (Quinlan, 1990; Quinlan & Cameron-Jones, 1993). Alternatively, forced simulation could be used in heuristic programs. For example, although forced simulation for programs with many recursive clauses is nondeterministic and hence potentially inefficient, one could introduce heuristics that would make the forced simulation efficient, at the cost of completeness.

A companion paper (Cohen, 1995) shows that the positive results of this paper are not likely to be improved: either eliminating the basecase oracle for the language above or learning two recursive clauses simultaneously is as hard as learning DNF, and learning $n$ linear recursive determinate clauses, one $n$-ary recursive determinate clause, or one linear recursive "$k$-local" clause is as hard as breaking certain cryptographic codes. With the positive results of this paper, these negative results establish the boundaries of learnability for recursive programs function-free in the pac-learnability model. These results thus not only give a prescription for building a formally justified system for learning recursive programs; taken together, they also provide upper bounds on what one can hope to achieve with an efficient, formally justified system that learns recursive programs from random examples alone.

## Appendix A. Additional Proofs

Theorem 1 states: Let $Dec = (p, a', R)$ be a declaration in $\in a\text{-}\mathcal{D}et\mathcal{DEC}^=$, let $n_r = \|R\|$, let $X_1, \ldots, X_{a'}$ be distinct variables, and define the clause $BOTTOM_d^*$ as follows:

$$BOTTOM_d^*(Dec) \equiv CONSTRAIN_{Dec}(DEEPEN_{Dec}^d(p(X_1, \ldots, X_{a'}) \leftarrow))$$

For any constants $d$ and $a$, the following are true:

- the size of $BOTTOM_d^*(Dec)$ is polynomial in $n_r$;

- every depth-$d$ clause that satisfies $Dec$ is equivalent to some subclause of $BOTTOM_d^*(Dec)$.

**Proof:** Let us first establish the polynomial bound on the size of $BOTTOM_d^*$. Let $C$ be a clause of size $n$. As the number of variables in $C$ is bounded by $an$, the size of the set $\mathcal{L}_D$ is bounded by

$$\underbrace{n_r}_{(\# \text{ modes})} \cdot \underbrace{(an)^{a-1}}_{(\# \text{ tuples of input variables})}$$

Thus for any clause $C$

$$\|DEEPEN_{Dec}(C)\| \leq n + (an)^{a-1}n_r \qquad (1)$$

By a similar argument

$$\|CONSTRAIN_{Dec}(C)\| \leq n + (an)^a n_r \qquad (2)$$

Since both of the functions $DEEPEN_{Dec}$ and $CONSTRAIN_{Dec}$ give outputs that are polynomially larger in size than their inputs, if follows that composing these functions a constant number of times, as was done in computing $BOTTOM_d^*$ for constant $d$, will also produce only a polynomial increase in the size.

Next, we wish to show that every depth-$d$ determinate clause $C$ that satisfies $Dec$ is equivalent to some subclause of $BOTTOM_d^*$. Let $C$ be some depth-$d$ determinate clause, and without loss of generality let us assume that no pair of literals $L_i$ and $L_j$ in the body of $C$ have the same mode, predicate symbol, and sequence of input variables.[6]

Given $C$, let us now define the substitution $\theta_C$ as follows:

1. Initially set

$$\theta_C \leftarrow \{X_1^* = X_1, \ldots, X_{a'}^* = X_{a'}\}$$

   where $X_1^*, \ldots, X_{a'}^*$ are the arguments to the head of $BOTTOM_d^*$ and $X_1, \ldots, X_{a'}$ are the arguments to the head of $C$.

   Notice that because the variables in the head of $BOTTOM_d^*$ are distinct, this mapping is well-defined.

2. Next, examine each of the literals in the body of $C$ in left-to-right order. For each literal $L$, let variables $T_1, \ldots T_k$ be its input variables. For each literal $L^*$ in the body $BOTTOM_d^*$ with the same mode and predicate symbol whose input variables $T_1^*, \ldots, T_k^*$ are such that $\forall i : 1 \leq i \leq r, T_j^* \theta_C = T_j$, modify $\theta_C$ as follows:

$$\theta_C \leftarrow \theta_C \cup \{U_1^* = U_1, \ldots, U_l^* = U_l\}$$

   where $U_1, \ldots, U_l$ are the output variables of $L$ and $U_1^*, \ldots, U_l^*$ are the output variables of $L^*$.

   Notice that because we assume that $C$ contains only one literal $L$ with a given predicate symbol and sequence of input variables, and because the output variables of literals $L^*$ in $BOTTOM_d^*$ are distinct, this mapping is again well-defined. It is also easy to verify (by induction on the length of $C$) that in executing this procedure some variable in $BOTTOM_d^*$ is always mapped to each input variable $T_i$, and that at least one $L^*$ meeting the requirements above exists. Thus the mapping $\theta_C$ is *onto* the variables appearing in $C$.[7]

Let $A^*$ be the head of $BOTTOM_d^*$, and consider the clause $C'$ which is defined as follows:

- The head of $C'$ is $A^*$.

- The body of $C'$ contains all literals $L^*$ from the body of $BOTTOM_d^*$ such that either

  - $L^* \theta_C$ is in the body of $C$

  - $L^*$ is the literal $equal(X_i^*, X_j^*)$ and $X_i^* \theta_C = X_j^* \theta_C$.

We claim that $C'$ is a subclause of $BOTTOM_d^*$ that is equivalent to $C$. Certainly $C'$ is a subclause of $BOTTOM_d^*$. One way to see that it is equivalent to $C$ is to consider the clause $\hat{C}$ and the substitution $\hat{\theta}_C$ which are generated as follows. Initially, let $\hat{C} = C'$ and let $\hat{\theta}_C = \theta_C$. Then, for every literal $L = equal(X_i^*, X_j^*)$ in the body of $\hat{C}$, delete $L$ from $\hat{C}$, and finally replace $\hat{C}$ with $\hat{C}\sigma_{ij}$ and replace $\hat{\theta}_C$ with $(\hat{\theta}_C)\sigma_{ij}$, where $\sigma_{ij}$ is the substitution $\{X_i^* = X_{ij}^*, X_j^* = X_{ij}^*\}$ and $X_{ij}$ is some new variable not previously appearing in $\hat{C}$. (Note: by $(\hat{\theta}_C)\sigma_{ij}$ we refer to the substitution formed by replacing every occurrence of $X_i$ or $X_j$ appearing in $\hat{\theta}_C$ with $X_{ij}$.) $\hat{C}$ is semantically equivalent to $C'$ because the operation described above is equivalent to simply resolving each possible $L$ in the body of $C'$ against the clause "$equal(X,X)\leftarrow$".

---

6. This assumption can be made without loss of generality since for a determinate clause $C$, the output variables of $L_i$ and $L_j$ will necessarily be bound to the same values, and hence $L_i$ or $L_j$ could be unified together and one of them deleted without changing the semantics of $C$.

7. Recall that a function $f : X \leftarrow Y$ is *onto* its range $Y$ if $\forall y \in Y \exists x \in X : f(x) = y$.

The following are now straightforward to verify:

- $\hat{\theta}_C$ is a one-to-one mapping.

  To see that this is true, notice that for every pair of assignments $X_i^* = Y$ and $X_j^* = Y$ in $\theta_C$ there must be a literal $equal(X_i^*, X_j^*)$ in $C'$. Hence following the process described above the assignments $X_i^* = Y$ and $X_j^* = Y$ in $\hat{\theta}_C$ would eventually be replaced with $X_{ij}^* = Y$ and $X_{ij}^* = Y$.

- $\hat{\theta}_C$ is onto the variables in $C$.

  Notice that $\theta_C$ was onto the variables in $C$, and for every assignment $X_i^* = Y$ in $\theta_C$ there is some assignment in $\hat{\theta}_C$ with a right-hand side of $Y$ (and this assignment is either of the form $X_i^* = Y$ or $X_{ij}^* = Y$). Thus $\hat{\theta}_C$ is also onto the variables in $C$.

- A literal $\hat{L}$ is in the body of $\hat{C}$ iff $\hat{L}\hat{\theta}_C$ is in the body of $C$.

  This follows from the definition of $C'$ and from the fact that for every literal $L^*$ from $C'$ that is not of the form $equal(X_i^*, X_j^*)$ there is a corresponding literal in $\hat{C}$.

Thus $\hat{C}$ is an alphabetic variant of $C$, and hence is equivalent to $C$. Since $\hat{C}$ is also equivalent to $C'$, it must be that $C'$ is equivalent to $C$, which proves our claim. ∎

## Acknowledgements

The author wishes to thank three anonymous JAIR reviewers for a number of useful suggestions on the presentation and technical content.

## References

Aha, D., Lapointe, S., Ling, C. X., & Matwin, S. (1994). Inverting implication with small training sets. In *Machine Learning: ECML-94* Catania, Italy. Springer-Verlag. Lecture Notes in Computer Science # 784.

Angluin, D. (1988). Queries and concept learning. *Machine Learning, 2*(4).

Angluin, D. (1989). Equivalence queries and approximate fingerprints. In *Proceedings of the 1989 Workshop on Computational Learning Theory* Santa Cruz, California.

Bergadano, F., & Gunetti, D. (1993). An interactive system to learn functional logic programs. In *Proceedings of the 13th International Joint Conference on Artificial Intelligence* Chambery, France.

Biermann, A. (1978). The inference of regular lisp programs from examples. *IEEE Transactions on Systems, Man and Cybernetics, 8*(8).

Cohen, W. W. (1993a). A pac-learning algorithm for a restricted class of recursive logic programs. In *Proceedings of the Tenth National Conference on Artificial Intelligence* Washington, D.C.

Cohen, W. W. (1993b). Pac-learning non-recursive Prolog clauses. To appear in *Artificial Intelligence*.

Cohen, W. W. (1993c). Rapid prototyping of ILP systems using explicit bias. In *Proceedings of the 1993 IJCAI Workshop on Inductive Logic Programming* Chambery, France.

Cohen, W. W. (1994). Pac-learning nondeterminate clauses. In *Proceedings of the Eleventh National Conference on Artificial Intelligence* Seattle, WA.

Cohen, W. W. (1995). Pac-learning recursive logic programs: negative results. *Journal of AI Research, 2*, 541–573.

De Raedt, L., & Bruynooghe, M. (1992). Interactive concept-learning and constructive induction by analogy. *Machine Learning, 8*(2).

De Raedt, L., & Džeroski, S. (1994). First-order $jk$-clausal theories are PAC-learnable. In Wrobel, S. (Ed.), *Proceedings of the Fourth International Workshop on Inductive Logic Programming* Bad Honnef/Bonn, Germany.

De Raedt, L., Lavrač, N., & Džeroski, S. (1993). Multiple predicate learning. In *Proceedings of the Third International Workshop on Inductive Logic Programming* Bled, Slovenia.

Džeroski, S., Muggleton, S., & Russell, S. (1992). Pac-learnability of determinate logic programs. In *Proceedings of the 1992 Workshop on Computational Learning Theory* Pittsburgh, Pennsylvania.

Frazier, M., & Page, C. D. (1993a). Learnability in inductive logic programming: Some basic results and techniques. In *Proceedings of the Tenth National Conference on Artificial Intelligence* Washington, D.C.

Frazier, M., & Page, C. D. (1993b). Learnability of recursive, non-determinate theories: Some basic results and techniques. In *Proceedings of the Third International Workshop on Inductive Logic Programming* Bled, Slovenia.

Gold, M. (1967). Language identification in the limit. *Information and Control, 10*.

Hirsh, H. (1992). Polynomial-time learning with version spaces. In *Proceedings of the Tenth National Conference on Artificial Intelligence* San Jose, California. MIT Press.

Idestam-Almquist, P. (1993). Generalization under implication by recursive anti-unification. In *Proceedings of the Ninth International Conference on Machine Learning* Amherst, Massachusetts. Morgan Kaufmann.

King, R. D., Muggleton, S., Lewis, R. A., & Sternberg, M. J. E. (1992). Drug design by machine learning: the use of inductive logic programming to model the structure-activity relationships of trimethoprim analogues binding to dihydrofolate reductase. *Proceedings of the National Academy of Science, 89.*

Lavrač, N., & Džeroski, S. (1992). Background knowledge and declarative bias in inductive concept learning. In Jantke, K. P. (Ed.), *Analogical and Inductive Inference: International Workshop AII'92.* Springer Verlag, Daghstuhl Castle, Germany. Lectures in Artificial Intelligence Series #642.

Ling, C. (1991). Inventing necessary theoretical terms in scientific discovery and inductive logic programming. Tech. rep. 301, University of Western Ontario.

Ling, C. (1992). Logic program synthesis from good examples. In *Inductive Logic Programming.* Academic Press.

Lloyd, J. W. (1987). *Foundations of Logic Programming: Second Edition.* Springer-Verlag.

Muggleton, S. (1994). Inverting implication. To appear in *Artificial Intelligence.*

Muggleton, S., & De Raedt, L. (1994). Inductive logic programming: Theory and methods. *Journal of Logic Programming, 19/20*(7), 629–679.

Muggleton, S., & Feng, C. (1992). Efficient induction of logic programs. In *Inductive Logic Programming.* Academic Press.

Muggleton, S., King, R. D., & Sternberg, M. J. E. (1992). Protein secondary structure prediction using logic-based machine learning. *Protein Engineering, 5*(7), 647–657.

Muggleton, S., & Page, C. D. (1994). A learnability model for universal representations. In Wrobel, S. (Ed.), *Proceedings of the Fourth International Workshop on Inductive Logic Programming* Bad Honnef/Bonn, Germany.

Muggleton, S. H. (Ed.). (1992). *Inductive Logic Programming.* Academic Press.

Nienhuys-Cheng, S., & Polman, M. (1994). Sample pac-learnability in model inference. In *Machine Learning: ECML-94* Catania, Italy. Springer-Verlag. Lecture notes in Computer Science # 784.

Page, C. D., & Frisch, A. M. (1992). Generalization and learnability: A study of constrained atoms. In *Inductive Logic Programming.* Academic Press.

Pazzani, M., & Kibler, D. (1992). The utility of knowledge in inductive learning. *Machine Learning, 9*(1).

Quinlan, J. R., & Cameron-Jones, R. M. (1993). FOIL: A midterm report. In Brazdil, P. B. (Ed.), *Machine Learning: ECML-93* Vienna, Austria. Springer-Verlag. Lecture notes in Computer Science # 667.

Quinlan, J. R. (1990). Learning logical definitions from relations. *Machine Learning, 5*(3).

Quinlan, J. R. (1991). Determinate literals in inductive logic programming. In *Proceedings of the Eighth International Workshop on Machine Learning* Ithaca, New York. Morgan Kaufmann.

Rouveirol, C. (1994). Flattening and saturation: two representation changes for generalization. *Machine Learning*, *14*(2).

Shapiro, E. (1982). *Algorithmic Program Debugging*. MIT Press.

Srinivasan, A., Muggleton, S. H., King, R. D., & Sternberg, M. J. E. (1994). Mutagenesis: ILP experiments in a non-determinate biological domain. In Wrobel, S. (Ed.), *Proceedings of the Fourth International Workshop on Inductive Logic Programming* Bad Honnef/Bonn, Germany.

Stahl, I., Tausend, B., & Wirth, R. (1993). Two methods for improving inductive logic programming. In *Proceedings of the 1993 European Conference on Machine Learning* Vienna, Austria.

Summers, P. D. (1977). A methodology for LISP program construction from examples. *Journal of the Association for Computing Machinery*, *24*(1), 161–175.

Valiant, L. G. (1984). A theory of the learnable. *Communications of the ACM*, *27*(11).

Zelle, J. M., & Mooney, R. J. (1994). Inducing deterministic Prolog parsers from treebanks: a machine learning approach. In *Proceedings of the Twelfth National Conference on Artificial Intelligence* Seattle, Washington. MIT Press.